\documentclass[preprint,12pt,authoryear]{elsarticle}

\usepackage{graphicx} 
\usepackage{amssymb}
\usepackage{amsmath}
\usepackage{subcaption}
\usepackage{bm}
\usepackage{booktabs}
\usepackage[ruled,vlined]{algorithm2e}  
\usepackage{ulem}     
\usepackage{xcolor}   
\usepackage{graphicx} 
\usepackage{makecell}
\usepackage{multirow} 
\usepackage{makecell} 
\usepackage{float}    

\newcommand{\ve}[1]{\mathbf{#1}}
\newcommand{\vg}[1]{\bm{#1}}
\newcommand{\ma}[1]{\mathbf{#1}}
\newcommand{\mg}[1]{\bm{#1}}
\newcommand{\T}{\mathsf{^T}}

\journal{Expert Systems with Applications}

\begin{document}

\begin{frontmatter}

\title{A Multivariate Statistical Framework for Detection, Classification and Pre-localization of Anomalies in Water Distribution Networks}
\date{August 2025}

\author[KhPI]{Oleg Melnikov\corref{cor1}} 
\ead{oleg.melnikov@khpi.edu.ua} 
\author[KhPI]{Yurii Dorofieiev}
\author[KhPI]{Yurii Shakhnovskiy}
\author[UofG]{Huy Truong}
\author[UofA]{Victoria Degeler\corref{cor1}}
\ead{v.o.degeler@uva.nl} 
\cortext[cor1]{Corresponding authors.}

\affiliation[KhPI]{organization={Institute of Computer Science and Information Technologies, National Technical University "Kharkiv Polytechnic Institute"},
            addressline={Kirpichova 2}, 
            city={Kharkiv},
            postcode={61002}, 
            country={Ukraine}}

\affiliation[UofA]{organization={Informatics Institute, University of Amsterdam},
            addressline={Science Park 900}, 
            city={Amsterdam},
            postcode={1098 XH}, 
            country={The Netherlands}}

\affiliation[UofG]{organization={Bernoulli Institute, University of Groningen},
            addressline={Nijenborgh 9}, 
            city={Groningen},
            postcode={9747 AG}, 
            country={The Netherlands}}           

\begin{abstract}
This paper presents a unified framework, for the detection, classification, and preliminary localization of anomalies in water distribution networks using multivariate statistical analysis.  The approach, termed SICAMS (Statistical Identification and Classification of Anomalies in Mahalanobis Space), processes heterogeneous pressure and flow sensor data through a  whitening transformation to eliminate spatial correlations among measurements. Based on the transformed data, the Hotelling's $T^2$ statistic is constructed, enabling the formulation of anomaly detection as a statistical hypothesis test of network conformity to normal operating conditions. It is shown that Hotelling's $T^2$ statistic can serve as an integral indicator of the overall “health” of the system, exhibiting correlation with total leakage volume, and thereby enabling approximate estimation of water losses via a regression model. 

A heuristic algorithm is developed to analyze the $T^2$ time series and classify detected anomalies into abrupt leaks, incipient leaks, and sensor malfunctions. Furthermore, a coarse leak localization method is proposed, which ranks sensors according to their statistical contribution and employs Laplacian interpolation to approximate the affected region within the network. 

Application of the proposed framework to the BattLeDIM L-Town benchmark dataset demonstrates high sensitivity and reliability in leak detection, maintaining robust performance even under multiple leaks. These capabilities make the method applicable to real-world operational environments without the need for a calibrated hydraulic model. 
\end{abstract}

\begin{highlights}
\item Statistical hypothesis testing as a unified framework for anomaly detection in water distribution networks.
\item Using Hotelling's $T^2$ statistic for leakage detection and size assessment.
\item Applicability in supervised and unsupervised settings with minimal training data.
\item Post-processing and visualization for anomaly classification and pre-localization.
\end{highlights}


\end{frontmatter}

\section{Introduction}
\label{sec:intro}

Water Distribution Networks (WDNs) are critical infrastructures that support the daily functioning of modern societies by providing clean water to consumers. These networks are complex and vulnerable to infrastructure failures, which can lead to substantial water losses. According to estimates reported in \cite{LW19}, more than 30\% of potable water is lost worldwide, approximately \$39 billion annually. This problem becomes more acute as world populations grow, while global warming and other environmental factors may lead to the depletion of available fresh water resources. 

A significant portion of water losses is attributed to leakages, 
which are commonly categorized as
background, unreported, and reported \citep{american2008water}. 
Background leaks are minor, persistent seepages, while reported leaks are major failures typically observed on the surface and reported by customers or authorities.
The greatest losses arise from unreported leaks that remain undetected for extended periods. Following \cite{Vrachimis2022}, leaks can also be classified by their temporal evolution as abrupt leaks, where the leakage rate remains constant over time, and incipient leaks, where the leakage rate gradually increases to a steady maximum. 


Leak detection in underground pressurized pipelines is challenging because of their limited accessibility for monitoring. Furthermore, network operation is influenced by multiple stochastic factors, including variations in nodal water demand, aging and clogging of the pipes, changes in pump and valve settings, and sensor malfunctions \citep{Mohan2023}.

All of these factors can lead to anomalies, i.e. deviations of the state of the network from the expected operational regime. In this context, three interrelated tasks are distinguished:
(i) anomaly detection, which identifies abnormal network behavior;
(ii) anomaly classification, which differentiates between leaks and sensor faults and identifies the type of leak; and
(iii) anomaly localization, which determines the specific pipe or junction associated with the anomaly.
Developing methods capable of efficiently and quickly addressing these tasks is crucial, as they help mitigate negative impacts and prevent further degradation of infrastructure.



    
    
    
It is known that any leak causes a pressure drop at its point of origin, propagating through the pipe network \citep{SILVA1996S491}. This makes pressure sensors a practical means of monitoring WDN conditions. 
Pressure sensors are often preferred to flow meters due to their lower cost; however, they 
provide only indirect indication of leaks.
Several additional factors further complicate leak detection and localization in real WDNs, including:

1.~Sparse sensor deployment. In practice, only a small fraction of the nodes (typically 3--7\%) are equipped with pressure sensors, significantly limiting spatial observability.

2.~Correlation between sensor readings. Due to the hydraulic connectivity of the network, the pressure drops caused by leaks propagate through the network, producing strong statistical dependencies between sensor outputs.

3.~Presence of undetected leaks. Persistent, low-intensity leaks can remain unnoticed and distort the statistical properties of the system, making it difficult to establish a reliable baseline or a “normal” operating state.

4.~Overlapping leakage events. New leaks may occur before existing ones are repaired, resulting in multiple simultaneous anomalies whose combined effects are difficult to disentangle.

These factors can cause false alarms, increase the detection delay, and impose strict requirements on the accuracy of hydraulic models. Despite these challenges, the past decade has seen extensive research devoted to developing methods for leak detection and localization in WDNs. An analysis of recent literature reviews \citep{Wan2022, Romero2023} suggests that data-driven approaches 
tend to perform best for anomaly detection, while hydraulic model-based methods 
are more effective for anomaly localization.

This study advances anomaly detection in WDNs by framing the task within a multivariate hypothesis testing framework.
Our approach, termed SICAMS (Statistical Identification and Classification of Anomalies in Mahalanobis Space)   
accounts for spatial inter-sensor correlations to improve the reliability of anomaly detection in WDNs. 
We demonstrate that Hotelling's $T^2$ statistic can serve as a robust global network health indicator. 
In addition, several post-processing procedures are proposed to classify detected anomalous events and to perform their rough localization.

The remainder of the paper is organized as follows. Section \ref{sec:lit} reviews the related literature. Section \ref{sec:matmet} presents our proposed framework and discusses details of its practical implementation. Section \ref{sec:data} describes the L-Town dataset used in this study. Section \ref{sec:results} addresses the validation of our methodology using this dataset as a reference. Section~\ref{sec:extensions} explores related issues, such as the assessment of water losses in the network and the feasibility of training without a dedicated dataset.   
Section \ref{sec:conc} concludes and discusses possible directions for future work.


\section{Related Works}
\label{sec:lit}
The importance of anomaly detection and localization in WDNs has led to a surge in research. A bibliometric analysis of the Scopus database by \cite{Farah2023}  identified over 600 publications on this topic between 2000 and 2023, with research activity accelerating markedly in the last decade. Numerous reviews \citep{Hu2021, Yussof2021, Negm2023, Romero2023, Nimri2023} have summarized the main methods and technologies developed to date.

Existing approaches are commonly classified as hardware- or software-based \citep{Romero2023}. Hardware-based methods are based on direct inspection using specialized field equipment, most commonly acoustic detection \citep{Liu2024}. Software-based methods, in contrast, process data collected from stationary sensors  via Supervisory Control and Data Acquisition (SCADA) systems. Although computationally intensive, they enable both snapshot and time-series detection and often outperform hardware-based techniques in operational monitoring.

Recent studies further classify leak detection and localization techniques according to their reliance on hydraulic modeling into (i) model-based methods, (ii) data-driven methods, and (iii) hybrid approaches combining both.

Model-based methods use hydraulic numerical models grounded in fluid mechanics to compute the hydraulic state of a WDN  while ensuring mass and energy conservation.  The relationship between  pipe flow and head loss  is described by the Bernoulli equation,  typically supplemented by empirical formulas  such as Darcy–Weisbach or Hazen–Williams to account for friction and other losses. Together with equations that govern the operation of pumps, valves, and other controlled components, these relations describe the water transport processes in pressurized pipe networks.

This approach underpins simulation tools such as the widely used EPANET ~\citep{Rossman2000},
where the WDN’s topology and configuration are defined in an input file that serves as the basis for simulations. Typically, such models require calibration of nodal demand patterns, pipe roughness coefficients, and other parameters to reproduce the observed network behavior accurately.  

The  fundamental principle of model-based anomaly detection  is to compare sensor readings: measured pressures, flows, and tank levels with simulated values under nominal operating conditions. Significant discrepancies indicate possible anomalies, such as leaks or other abnormal operating states.


According to the review by \cite{Hu2021}, model-based approaches can be divided into sensitivity matrix methods, optimization and calibration approaches, and error-domain model falsification methods. \cite{Marzola2022} conducted an empirical comparison of different techniques under varying sensor densities, calibration accuracies, and data aggregation intervals, showing that both the quality of model calibration and spatial sensor placements significantly affect anomaly detection performance.

Other representative model-based approaches include 
Kalman filtering \citep{Preis2011}, residual clustering \citep{LiandXin2020}, and 
exploiting hydraulic duality \citep{Stef2022}. 
These methods differ in data requirements, computational complexity, and robustness to modeling errors, providing distinct trade-offs for practical monitoring.

A main limitation of model-based approaches is the difficulty of creating a hydraulic model that accurately reflects the real network, whose characteristics change during operation. Parameter calibration only partially compensates for these discrepancies, and even minor modeling errors may significantly affect leak detection and localization accuracy.

As an alternative, data-driven methods identify anomalies directly from SCADA data on pressure, flow, and consumption without the need for a calibrated hydraulic model. 

\cite{Wan2022} classify them into three major categories: statistical process control (SPC) methods, prediction and classification approaches, and feature extraction and clustering techniques.

Prediction and classification methods are the most prevalent in the literature. They employ machine learning models, typically based on artificial neural networks, that are trained on historical data representing normal operating conditions. Anomalies are detected by analyzing the deviations between the observed and predicted system variables.

Feature-based and clustering methods extract informative attributes from pressure and flow data and classify them using clustering algorithms such as k-means \citep{Min2022} or supervised classifiers, including support vector machines \citep{Cai2022} and random forests \citep{Shen2022}. \cite{Nimri2023} concluded that data-driven methods  generally outperform model-based ones in handling uncertainty during leak detection; however, their performance depends on the availability of large, high-quality labeled datasets, which are not always accessible.

The use of graph neural networks (GNNs)  represents another promising direction, as these models explicitly incorporate network topology and capture both local and global dependencies.  
As a recent example, \cite{Truong2024} proposed a residual GNN architecture trained with randomly placed sensors, enabling accurate pressure reconstruction at unmonitored nodes.

Statistical methods require neither predictive models nor classification architectures and are considered conceptually simple but effective in identifying anomalous WDN states. They are typically based on control charts that detect outliers in monitored data using thresholds computed from historical statistics. When real-time SCADA data exceed these thresholds, they are flagged as anomalous.

The most widely used univariate statistical methods include the Shewhart chart \citep{Loureiro2016}, Western Electric rules \citep{Ahn2019}, and Exponentially Weighted Moving Average control charts \citep{Jung2015}. However, these methods do not account for correlations among multiple sensor readings. Consequently, recent studies have focused on multivariate statistical methods, which enhance the efficiency and reliability of anomaly detection.
For example, \cite{Barros2025} stratified the pressure sensor data, classified the network nodes using the PageRank algorithm, and detected anomalies using z-scores and interquartile range analysis.





\cite{ChenYang2022} proposed a multivariate statistical leakage model based on Random Matrix Theory (RMT). Their method includes three stages: (i) forecasting short-term water demand using the Elman neural network; (ii) real-time leak localization based on a hydraulically calibrated model using genetic algorithms; and (iii) estimation of leak size through RMT. The main drawbacks of this approach are a relatively high false-alarm rate (since the demand forecast does not account for seasonal variations) and the use of fixed leak sizes during training, which hinders the detection of gradually developing leaks.

\cite{Xie2024} introduced a multi-point leakage prediction framework using Bayesian inference, which models the probability of leaks of various magnitudes at different network locations while accounting for data uncertainty. The limitations of this method include the need for probabilistic descriptions of model and measurement errors and the dependence on data quality, as systematic sensor biases can distort probabilistic estimates.

Both of these methods can be regarded as hybrid approaches, as they integrate data-driven analysis with hydraulic simulations using EPANET, thus requiring a calibrated network model.

The diversity of 
leak detection methodologies highlighted the need for a standardized platform 
for performance comparison. To this end, the Battle of the Leakage Detection and Isolation Methods (BattLeDIM) competition was organized in 2020. Its organizers provided a common benchmark WDN, known as L-Town, to evaluate leak detection methods. Eighteen teams from academia and industry participated, and their results were reviewed by \cite{Vrachimis2022}. The evaluation considered both detection and localization accuracy as well as economic efficiency in terms of water loss and localization cost. Even the best solutions achieved only ~50\% of the theoretical optimum, underscoring the complexity of the problem and the need for continued research.


A relatively underexplored topic in the literature is anomaly classification, which involves distinguishing between abrupt and incipient leaks, sensor faults, and atypical demand patterns. To our knowledge, only one study \citep{CaiGaoXu2022} proposed a method that performs both detection and classification of anomalies in WDNs using continuously updated water consumption data. The method combines hourly demand forecasting models with feature extraction techniques and convolutional neural networks. These models are self-learning, operate in real time, and do not require a hydraulic network model. However, the authors used ad hoc features for anomaly classification, which limits the general applicability of the approach. Consequently, the development of a unified statistical framework capable of both detecting and classifying anomalies in WDN operation remains an open challenge.

In general, existing data-driven methods cannot fully eliminate the effects of strong spatial correlations among sensor readings, while model-based approaches remain constrained by the need for accurate and frequently recalibrated hydraulic models.

Therefore, the objective of this research is to develop a methodology that (i) enhances the robustness of data-driven approaches by mitigating spatial correlations among sensors, and (ii) avoids the dependence on precise hydraulic model calibration characteristic of model-based methods.  

The proposed framework employs multivariate statistical techniques commonly used in quality control for the simultaneous monitoring of correlated variables and for detecting changes in the overall system state \citep{MahExample1, MahExample2}. 
However, they have seen limited application in the context of WDNs. To our knowledge, the only closely related work in this domain is the PCA-based burst detection method by  \cite{palau2012burst}. In that approach, hourly water flows are projected onto a principal-component subspace, and anomalies are identified using Hotelling’s $T^2$ statistic. 

Our framework differs in several important aspects. First, PCA is used primarily for dimensionality reduction and requires separate statistical models for different demand patterns, whereas our method employs a whitening transformation that fully decorrelates sensor measurements without relying on PCA or demand pattern segmentation. Second, \cite{palau2012burst} focus mainly on detecting abrupt bursts under relatively stable conditions, while our framework handles heterogeneous pressure and flow data, distinguishes between anomaly types, and provides coarse localization without a hydraulic model. Third, we demonstrate that the $T^2$ statistic can serve as a global health indicator of the network, retaining discriminative power even in the presence of multiple leaks -- a capability not addressed in earlier studies.

To summarize, our paper makes the following contributions.

1.~We introduce a multivariate statistical approach to anomaly detection in WDNs that employs a whitening transformation to decorrelate pressure and flow measurements, thereby enhancing the robustness and sensitivity of the detection process. The resulting transformed data are summarized by Hotelling’s $T^2$  statistic, which tracks the network state and serves as a basis for statistical tests. The methodology is applicable under both supervised and unsupervised conditions and does not require large labeled datasets.




2.~A heuristic algorithm is proposed to classify structural change points in the $T^2$ statistic obtained via the PELT algorithm, allowing discrimination between abrupt and incipient leaks.

3.~A procedure is outlined to distinguish genuine leak events from anomalies caused by sensors with systematic measurement errors.

4.~A coarse localization method is developed that identifies the most responsive sensors and applies Laplacian interpolation to estimate the spatial extent of the affected area within the network.  Unlike most detection methods, this approach does not require a calibrated hydraulic model. 

5.~A leak size estimation model is proposed based on a regression on the Wilson–Hilferty transformed moving average of the $T^2$ statistic.


\section{Methodology}
\label{sec:matmet}
This section presents a detailed description of the proposed methodology.  
We begin by formalizing the leak detection problem and introducing the notation that we use throughout this paper in Section \ref{sec:Problem}. Next, the overall mathematical framework for anomaly detection based on the Mahalanobis metrics is described in Section \ref{sec:overview}. We then discuss details of the practical implementation of the methodology in Section \ref{sec:details}. Section~\ref{sec:pelt} describes handling multiple leakages. Finally, Section~\ref{sec:classloc} addresses classification and pre-localization of anomalies.

\subsection{Problem Statement}
\label{sec:Problem}
The structural elements of a WDN can be broadly categorized into controlled and uncontrolled components. Controlled elements include pumps and valves, whose operational parameters can be adjusted. Uncontrolled components consist of pipes, storage tanks, and junction nodes, i.e., points in the network where water flows merge or split.

The structure of a WDN is typically represented as an undirected graph $G = (V, E)$, where $V$ denotes the set of nodes (e.g., junctions and tanks), and $E$ the set of edges (pipes). Some nodes and edges are instrumented with sensors of various types, such as pressure sensors, flow meters, water level sensors (for tanks), and automated meter reading (AMR) devices. 

The hydraulic state of the network at time $t$ is only partially observable due to the limited sensor coverage. We denote this observable state by a vector $ \ve{x}_t \in \mathbb{R}^s$, where $s$ is the number of sensors. This vector can be partitioned as
$
\ve{x}_t = \left( \ve{p}_t, \ve{f}_t, \ve{w}_t \right),
$
where $\ve{p}_t$ represents the pressure sensor readings in a subset of nodes $V_p \subset V$, $\ve{f}_t$ denotes the flow sensor measurements in a subset of pipes $E_f \subset E$, and $\ve{w}_t$ corresponds to the water level measurements in tanks. These data are collected in real time at regular intervals by the SCADA system.

Let $S_t$ be a binary variable indicating the system status at time $t$ (0 for normal operations, 1 for anomalous). We define an anomaly detection function as a mapping 
\begin{equation}
    S_t = \varphi\left(\ve{x}_{t - l}, \dots, \ve{x}_{t - 1}, \ve{x}_t ; S_{t-1}\right),
    \label{eq:def1}
\end{equation}
where $l$ is the depth of the observation window. A particularly simple and desirable case arises when $l=0$, in which case the detection rule becomes Markovian. In practice, however,  the detection algorithm may rely on smoothing the data over a moving window to distinguish persistent deviations from transient fluctuations and to reduce false positives. 

If an anomaly occurs at time $\tau$ (with $S_{\tau-1}=0$), the detection delay may be defined as 
\begin{equation}
    \Delta t = \min\{t \ge \tau\,|\,S_t=1\}-\tau.
    \label{eq:Delta}
\end{equation}
 
This detection delay, together with the confusion matrix, provides a basis for evaluating the performance of a detection rule.

Problem~\eqref{eq:def1} can be addressed within either a supervised or an unsupervised learning framework. The supervised setting assumes the availability of a reference set of time indices $\mathcal{T}$ during which the system is known to have operated normally. In the unsupervised setting, no such prior knowledge is assumed, and anomalies must be inferred from the structure of the data alone.

Most existing approaches to anomaly detection in WDNs assume the presence of a single leak. This assumption is often violated in real-world scenarios. Several recent studies have demonstrated the feasibility of multi-leak detection \citep{Daniel2022, Stef2022, Wang2022}. Therefore, there is a need to expand the formulation of the anomaly detection problem to take into account multiple leaks.

\newcommand{\Ntrue}[1]{N_{#1}}
\newcommand{\Nest}[1]{\hat{N}_{#1}}

Let $\Ntrue{t}$ denote the true number of active leaks in the WDN at time $t$, and let $\Nest{t}$ be its estimated value. A multiple-leak detection rule can then be formulated as 
\begin{equation} 
    \left\{ \begin{array}{l}
        \Nest{t}  = \Nest{t-1}  + \xi _t,  \\ 
        \xi _t  = \psi \left(\ve{x}_{t - l} , \ldots ,\ve{x}_{t - 1} ,\ve{x}_t , \Nest{t-1} \right),  
    \end{array} \right.
    \label{eq:mdrule}
\end{equation}
where 
$$
\xi_t = \left\{
\begin{array}{rl}
+1 & \text{if a new leak is declared,} \\
\ \,0 & \text{if there is no change in status,} \\
-1 & \text{if an existing leak is considered resolved.}
\end{array}
\right.
$$

In the supervised learning environment, the performance of a multi-leak detection rule can be evaluated using a suitable distance metric $d(\Ntrue{t},\Nest{t})$. For instance, the time integral 

\begin{equation} 
    M=\int_{0}^{T} \lvert{\Ntrue{t}-\Nest{t}}\rvert{dt}.
    \label{eq:metrics}
\end{equation}
can be interpreted as the total number of leak-hours missed by a particular estimate $\Nest{t}$ over the observation period $[0,T]$. For an ideal estimate, $\Nest{t}=\Ntrue{t}$ and $M=0$.

Anomaly localization consists in identifying the set of pipes or nodes responsible for the occurrence of an anomaly. In a sufficiently general setting, this task can be formulated as the construction of a scoring vector with non-negative components $l(e), e \in E$ (or $l(v), v \in V)$, where a larger value indicates a higher likelihood of an anomaly (without necessarily implying a strict probabilistic interpretation).

If the scoring function is concentrated on a specific edge or node, localization can be considered precise; otherwise, it should be regarded as approximate pre-localization.

\subsection{General Framework}
\label{sec:overview}
Let $\ve{x} \in \mathbb{R}^s$ be a random column vector with mean $\vg{\mu}$ and a non-singular covariance matrix $\mg{\Sigma}$. Applying a whitening transformation 
\begin{equation}
    \ve{z}=\ma{W}(\ve{x}-\vg{\mu}), \text{ where } \ma{W}\T\ma{W}=\mg{\Sigma}^{-1},
    \label{eq:wht}
\end{equation}
 produces a white noise vector $\ve{z}$ with zero mean and unit diagonal covariance \citep{whitening2018}.  

 If, in addition, $\ve{x}$ is a multivariate normal vector $\ve{x} \sim \mathcal{N}(\vg{\mu},\mg{\Sigma})$, then $\ve{z} \sim \mathcal{N}(\ve{0},\ma{I})$, where $\ma{I}$ is the unit matrix. Then the squared Mahalanobis norm of $\ve{z}$ follows a chi-squared distribution with $s$ degrees of freedom:
\begin{equation}
    Z^2 = \|\ve{z}\|^2 = (\ve{x} - \vg{\mu})\T \mg{\Sigma}^{-1} (\ve{x} - \vg{\mu})\sim \chi^2(s).
    \label{eq:mahdis}
\end{equation}

Thus, in the context of the anomaly detection problem, if sensor measurements $\ve{x}$ obtained under normal operating conditions can be adequately modeled by a multivariate normal distribution  $\mathcal{N}(\vg{\mu},\mg{\Sigma})$, then the statistic \eqref{eq:mahdis} can serve as an indicator of the overall state of the network. In the presence of a leak or other abnormal condition, the vector $\ve{x}$ will deviate from its expected behavior, causing the test statistic $Z^2$ to increase and potentially exceed a predefined critical threshold $\chi^2_{1-\alpha}(s)$, where $\alpha$ is the significance level. This forms the basis of a statistical hypothesis test:

\begin{itemize}
    \item $H_0$: The system is operating normally, $\ve{x} \sim \mathcal{N}(\vg{\mu}, \mg{\Sigma})$
    \item $H_1$: An anomaly is present, and $\ve{x} \not\sim \mathcal{N}(\vg{\mu}, \mg{\Sigma})$
\end{itemize}

 As the components of $\ve{z}$ are independent standard normal variables, the same framework can be applied not only to the entire system but also to its components and individual sensors. For any subset of indices $A \subseteq \{1,2,\ldots,s\}$, the corresponding partial sum of squares will also follow a chi-squared distribution, but with $|A|$ degrees of freedom: 
 \begin{equation}
    Z^2_A = \sum_{i \in A} z_i^2 \sim \chi^2(|A|) \qquad \text{under } H_0,
    \label{eq:z2}
\end{equation}
where $|A|$ denotes the number of elements in $A$.

This permits testing for anomalies at various spatial resolutions. In particular, localizing the increase in Mahalanobis distance to a specific subset of sensors provides a natural and interpretable mechanism for preliminary leak localization. Thus, the suggested statistical framework supports both detection and coarse localization within the network.

In practice, exact values of the mean vector $\vg{\mu}$ and covariance matrix $\mg{\Sigma}$ are rarely known. If the whitening transformation is performed using a mean vector and covariance matrix estimated from a sample of $n$ observations $ \ve{x}^{(1)}, \ldots, \ve{x}^{(n)} \in \mathbb{R}^s $, then the resulting sum of squares will be more accurately described by 
Hotelling's $T^2$ distribution \citep{mva2015}.

Let
\begin{equation}
    \bar{\ve{x}} = \frac{1}{n} \sum_{i=1}^n \ve{x}^{(i)}, \qquad
\ma{S} = \frac{1}{n-1} \sum_{i=1}^n (\ve{x}^{(i)} - \bar{\ve{x}})(\ve{x}^{(i)} - \bar{\ve{x}})\T
\label{eq:estimates}
\end{equation}
denote the sample mean vector and the sample covariance matrix, respectively. Provided $n>s$, 
the Mahalanobis distance computed with estimated parameters
\begin{equation}
    T^2 = (\ve{x} - \bar{\ve{x}})\T \ma{S}^{-1} (\ve{x} - \bar{\ve{x}})
    \label{eq:teststat}
\end{equation}
follows Hotelling's $T^2$ distribution with $s$ and $n$ degrees of freedom.

Statistic \eqref{eq:teststat} can be converted to the more common Fisher distribution by applying the following scaling: 
\begin{equation}
    T^2_F=\frac{n - s}{s(n - 1)} T^2 \sim F(s, n - s)
    \label{eq:t2f}
\end{equation}
where $ F(d_1, d_2) $ denotes the Fisher distribution with $ d_1 $ and $ d_2 $ degrees of freedom.

As the sample size increases, the Hotelling's distribution asymptotically converges to the chi-squared distribution.

\subsection{Implementation Details}
\label{sec:details}

\begin{figure}
    \centering
    \includegraphics[width=0.95\linewidth]{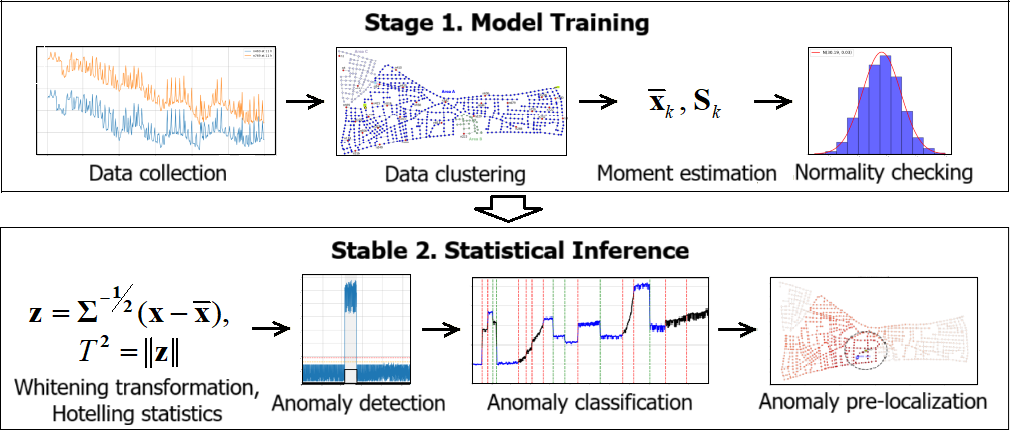}
    \caption{Flowchart of the proposed method} 
    \label{fig:flowchart}
\end{figure}

Practical implementation of the proposed framework involves two major stages, which are illustrated by the flowchart in Fig.~\ref{fig:flowchart}.

1.~Training the model using data corresponding to normal operational conditions.

2.~Applying the model to operational or historical data for anomaly detection.

\textbf{Stage 1} entails estimating the statistical properties of the sensor measurement vector under nominal (anomaly-free) conditions.

Sensor data exhibit substantial spatial and temporal variability. To account for this, the estimation procedure comprises the following steps:

1a)~if applicable, partitioning the network into spatial clusters that can be treated as approximately independent;

1b)~grouping time points into temporal clusters;

1c)~estimating the mean vectors and covariance matrices for all spatial and temporal clusters.

The output of \textbf{Step 1a} is a partitioning of the sensor index set ${1,2,\ldots,s}$ into several spatial clusters. The primary objective of this step is to reduce the dimensionality of the problem.
The partitioning may rely on the WDN physical layout or be derived using clustering algorithms. An overview of methods for WDN partitioning can be found in \cite{Bui2020}. Since the subsequent procedures are applied identically to each spatial cluster, below we will focus on a single representative cluster to simplify notation.

\textbf{Step 1b} involves selecting a statistical model to represent the temporal structure of the data vector. 
To this end, we group the data into temporal clusters. More formally, let $ t = 1, 2, \ldots, T $ denote discrete time points over the observation period. We define a mapping $\tau: \{1, 2, \ldots, T\} \rightarrow \mathcal{C}$, which assigns each time point $t$ to a corresponding temporal cluster  $\tau(t) \in \mathcal{C} $. This mapping can vary for different spatial clusters.

Temporal clusters are identified based on both calendar structure (e.g., hour of day, day of week, month) and statistical similarity between multivariate observations $\ve{x}_t \in \mathbb{R}^s$. The resulting segmentation should capture both periodic regularities and intrinsic time patterns present in the data. 

The components of the vectors $\ve{x}$ in each temporal cluster should be assessed for normality. This check is less critical when the dimensionality of $\ve{x}$ is high, as in such cases the expression in equation~\eqref{eq:z2} will converge to a normal distribution due to the central limit theorem, even if some individual components deviate from normality.

It is also possible to adopt a more structured approach to covariance matrix estimation, particularly when available data are limited. For instance, if the correlation matrix $\ma{K}$ remains relatively stable over time, 
the covariance matrix can be modeled as
\begin{equation}
    \ma{S}_{ij}(t)=\sigma_i(t)\sigma_j(t)\ma{K}_{ij},\quad i,j = {1,\ldots,S},
\end{equation}
where $\sigma_i(t)$ denotes the time-varying standard deviation of the $i^{th}$ component of the measurement vector.

\textbf{Step 1c} involves moment estimation for sensor data, separately for each temporal cluster. The estimation approach depends on whether the data originates from a hydraulic model or from SCADA time series.

In the case of hydraulic simulation systems (e.g., EPANET), computations are deterministic, and sensor values computed at different time steps can be taken as their mean vectors $\vg{\mu}_t$. To compute covariance matrices, the model must be randomized by introducing uncertainty in parameters such as base demands, demand time patterns, or pipe roughness. Let $ \ve{x}^{(i)}_t $ denote the vector of sensor values at time $t$ in the $i^{th}$ simulation run, $i=1,\ldots,N$. The covariance matrix for temporal cluster $k$ is then estimated as
\begin{equation}
    \ma{S}_k = \frac{1}{N\cdot|T_k|} \sum_{i=1}^N \sum_{t \in T_k} (\ve{x}^{(i)}_t - {\vg{\mu}_t})(\ve{x}^{(i)}_t - {\vg{\mu}_t})\T, 
    \label{eq:estcm_sim}
\end{equation}
where $T_k=\{t: \tau(t)=k\}$.

Equation~\eqref{eq:estcm_sim} enables the estimation of the statistical properties of $\ve{x}_t$, with accuracy and detail constrained only by available computational resources. 

However, the simulation parameters must be selected so that they reflect the variance of the real data obtained from the SCADA system.

When using time series data, the estimation relies on a single historical realization. This imposes constraints on the number and size of temporal clusters, as accurate estimation requires that $|T_k| \gg s$. Furthermore, selecting time periods corresponding to normal system behavior may be challenging due to the possible presence of undetected anomalies. These issues will be discussed in greater detail in Section~\ref{sec:results}.

\textbf{Stage 2} of the framework consists in applying the whitening transformation \eqref{eq:wht} to operational or historical data $\{\ve{y}_t\}$, which have the same format as the training data. For each observation, the corresponding temporal cluster is identified, and the associated mean vector and covariance matrix are used for normalization. The squared Mahalanobis distance is then computed according to Equation~\eqref{eq:teststat} and rescaled to the $T^2_F$ statistic using Equation~\eqref{eq:t2f}. The resulting value is then compared to a predefined critical threshold.

The choice of the whitening matrix $\ma{W}$ in Equation~\eqref{eq:wht} is not unique. We have used Cholesky whitening ($\ma{W}=\ma{L}\T$, where $\ma{L}$ is the Cholesky decomposition of $\mg{\Sigma}$)  because of its computational efficiency. Other possible choices are Mahalanobis whitening $\ma{W}=\mg{\Sigma} ^{-1/2}$, or the eigenvalue decomposition of $\mg{\Sigma}$ \citep{whitening2018}.

To improve robustness against noise and reduce false positives, we employ a hysteresis-based decision rule. An anomaly is considered detected if the moving average of the $T^2_F$ statistic over a specified period of time exceeds the threshold $\theta_1$. Conversely, the anomaly is considered resolved once it returns to the baseline level $\theta_0$. 

From Equation \eqref{eq:t2f}, the critical value $\theta_1$ for cluster $k$ for a desired significance level $\alpha$ is given by 
\begin{equation}
    \theta_1(\alpha,k)=F_{1-\alpha}(s,n_k-s),
    \label{eq:theta1}
\end{equation}
where $n_k$ is the number of observations for the $k^{th}$ temporal cluster, and $F_q(d_1,d_2)$ is the $q^{th}$ quantile of the Fisher distribution with $(d_1,d_2)$ degrees of freedom. 

For $\theta_0$, we use the expected value of the $T^2_F$ statistic. By the properties of the Fisher distribution, it follows from \eqref{eq:teststat} that, provided $n_k>s+2$,
\begin{equation}
    \theta_0(k)=\frac{n_k-s}{n_k-s-2}.
    \label{eq:theta0}
\end{equation}

If $n_k \gg s$, it may be more convenient to work with a statistic $T^2$ \eqref{eq:teststat} directly, as it would converge in distribution towards $\chi^2(s)$. 

The overall detection scheme is summarized in Algorithm 1.

\begin{algorithm}[t]
\caption{Anomaly Detection Using Whitening Transformation} 
\KwIn{
Observation sequence $\{\ve{y}_t\}_{t=1}^T$; 
temporal clustering map $\tau(t)$\; 
trained means $\vg{\mu}_k$ and 
covariance matrices $\mg{\Sigma}_k$\
for all clusters $k \in \mathcal{C}$\; 
detection window size $L$; 
critical threshold $\theta_1$; 
baseline threshold $\theta_0$
}
\KwOut{Anomaly detection status for each time point}

Initialize status $\text{AnomalyDetected} \gets \texttt{False}$\;
Initialize buffer $B \gets \emptyset$\;

\For{$t \gets 1$ \KwTo $T$}{
    Determine cluster $k \gets \tau(t)$\;
    Compute whitened vector: $\ve{z}_t = \ma{W}_k(\ve{y}_t - \vg{\mu}_k)$\;
    Compute squared norm: $T^2_t = \|\ve{z}_t\|^2$\;
    Compute $T^2_{F,t}$ according to equation \eqref{eq:t2f}\;
    Append $T^2_{F,t}$ to buffer $B$ (keep most recent $L$ values)\;
    Compute $\bar{T}^2_F(B) = \frac{1}{B}\sum_{i \in B} T^2_{F,i}$\;
    \eIf{\textnormal{AnomalyDetected == False}}{
        \If{$\bar{T}^2_F(B) > \theta_1$}{
            \texttt{AnomalyDetected} $\gets$ \texttt{True}\;
        }
    }{
        \If{$\bar{T}^2_F(B) < \theta_0$}{
            \texttt{AnomalyDetected} $\gets$ \texttt{False}\;
        }
    }
}
\end{algorithm}

If an anomaly is detected, several subsequent steps may be required.

1.~Determination of anomaly type – distinguishing between abrupt and incipient leaks, sensor malfunctions, or other system disruptions.

2.~Quantification of anomaly severity – for example, estimating the leakage magnitude or flow deviation.

3.~Localization of the anomaly source within the hydraulic network.

Possible approaches to addressing these problems will be discussed below.

\subsection{Handling Multiple Leakages}
\label{sec:pelt}
Anomaly recognition becomes more challenging when the system does not operate under normal conditions. In such cases, it may be necessary not only to confirm the resolution of a previously detected anomaly, but also to identify the emergence of a new anomaly superimposed over existing ones.

To address this problem, two alternative strategies may be employed.

1.~Retraining the model over a different time interval that incorporates background leakage as part of the system’s new operational status quo. 

2.~Using $T^2_F$ as a tracking statistic combined with change-point detection. In this strategy, the $T^2_F$ statistic is treated as a proxy for the network's operational state. Structural change-point (SCP) detection methods, such as PELT \citep{pelt2012}, can then be applied to identify and classify potential leakage events. 

Following the automatic detection of SCPs via PELT,  two additional post-processing tasks arise: 

(1) to classify the detected SCPs as \textit{‘anomaly start’} or \textit{‘anomaly end’}; 

(2) to remove spurious SCPs that may arise due to the presence of an 
incipient leak.

To address these tasks, the following heuristic algorithm was developed:

1.~The intervals between consecutive SCPs are bisected. For each resulting pair of sub-intervals, the mean values of the $T^2_F$ statistic are computed. A paired Student’s $t$-test is then used to assess whether the difference in means is statistically significant.

2.~If the sub-intervals under consideration are not separated by an SCP and their mean values differ significantly, the overall interval is considered \textit{increasing}. If the means do not differ significantly, the interval is labeled as \textit{stable}. 

3.~If the sub-intervals lie on opposite sides of an SCP and the means differ significantly, the SCP is classified based on the direction of the mean change: it is labeled as \textit{‘anomaly start’} if the mean increases, or \textit{‘anomaly end’} if the mean decreases.

4.~If two adjacent intervals, separated by an SCP, are both labeled as increasing, a linear regression model is fitted to each interval. The slopes of the two regression lines are then compared. If the absolute difference between the slopes is statistically significant, the SCP is considered redundant and removed from further analysis.

A more formal representation of this scheme is given in Algorithm 2. 

\begin{algorithm}[h]
\caption{Post-processing of Structural Change Points (SCPs)}
\KwIn{$T^2_F$ statistic moving average series; detected SCPs from PELT; significance level $\alpha$; slope threshold $\delta$}
\KwOut{Classified SCP:} 

\ForEach{interval $I$ between consecutive SCPs}{
    Bisect $I$ into two subintervals $I_1$ and $I_2$\;
    Compute $\bar{T}_1$ and $\bar{T}_2$ = mean $T^2_F$ values for $I_1$ and $I_2$\;
    $p \gets p$-value from paired Student's $t$-test for $\bar{T}_1$ vs $\bar{T}_2$\;
    
    \eIf{$p$ $< \alpha$}{
        Mark $I$ as \texttt{increasing}\;
    }{
        Mark $I$ as \texttt{stable}\;
    }
}

\ForEach{SCP between intervals $I_a$ and $I_b$}{
    \If{$p < \alpha$ for $\bar{T}_a$ vs $\bar{T}_b$}{
        \eIf{$\bar{T}_b > \bar{T}_a$}{
            Label SCP as \texttt{anomaly start}\;
        }{
            Label SCP as \texttt{anomaly end}\;
        }
    }
}

\ForEach{pair of adjacent \texttt{increasing} intervals $I_a$, $I_b$ separated by SCP}{
    Fit linear regression to $I_a$ and $I_b$; obtain slopes $m_a$ and $m_b$\;
    \If{$|m_a - m_b| < \delta$}{
        Remove SCP as redundant\;
    }
}
\end{algorithm}

\subsection{Classification and Pre-localization of Anomalies}
\label{sec:classloc}
Although leaks and pipe bursts represent the most common anomalies in WDNs (and pose the greatest risk to infrastructure integrity), they are not the only anomaly type. Other possible events include \citep{jian2022}:

$\bullet$ blockages and deposits;

$\bullet$ pump or valve failures (e.g., pumps operating outside schedule, stuck gate valves, malfunctioning control valves);

$\bullet$ sensor malfunctions;

$\bullet$ unauthorized water withdrawal.

SICAMS detects anomalies in general rather than leaks specifically. This distinction arises from the definition of the Hotelling's $T^2$ statistic, which treats positive and negative deviations symmetrically. In contrast, leaks typically manifest as one-sided changes, such as a systematic pressure drop. Thus, the subsequent and non-trivial task of classifying the type of detected anomaly must be addressed.

This problem was already partially considered in Section~\ref{sec:pelt}, where leaks were classified into incipient and abrupt. More generally, the development of classification rules requires the identification of discriminative features that distinguish one anomaly type from another. For example, in the case of a faulty sensor, the anomaly effect remains localized to a single measurement point, whereas a leak induces consistent changes across multiple sensors due to the strong connectivity of the network. Unauthorized water withdrawal resembles a leak in its hydraulic impact, but unlike leaks, it occurs intermittently and only during specific time windows.

Useful auxiliary statistics for anomaly classification include:

$\bullet$ individual sensor readings,

$\bullet$ the number of sensors exceeding their operational range,

$\bullet$ temporal persistence and coherence of anomalies in the time series,

$\bullet$ spatial consistency of deviations across hydraulically related nodes.

An illustrative case of anomaly classification will be considered in Section~\ref{sec:classify}.


Leak localization can be considered in two main variants: precise localization and coarse localization. Precise localization assumes the selection of the leak location according to some optimality criterion. In the studies cited above, e.g. \cite{Marzola2022}, precise localization was performed using a hydraulic model by simulating a leak in every pipe of the network, obtaining the corresponding sensor readings, and comparing them to the actual measurements. The most likely leak location is then chosen according to the criterion of minimal distance between these vectors. 


In many practical situations, however, a coarser leak localization is sufficient. When a repair crew is dispatched, acoustic loggers with a long detection range are typically employed to pinpoint the leak. For example, in the BattLeDIM competition, a leak was considered detected if the predicted location was within a radius of 300 m from the actual leak site. Thus, it is often enough to identify a subset of pipes likely to contain the leak, represented as a spatial region on the map. Following \cite{Mazzoni2024}, we refer to this as pre-localization.

Pre-localization can be performed in two conceptually different ways:

    (1) by partitioning the network into spatial clusters that can be analyzed separately to narrow the search down to a single cluster;
    
    (2) by identifying the sensors with the most anomalous readings and defining the effective coverage area of these sensors, which is then treated as the candidate search zone.

For our method, the natural approach is to identify the sensors that contributed the most to the increase in the test statistic. Hotelling's $T^2$  statistic given in Equation \eqref{eq:teststat} can be decomposed into sensor-wise contributions as  
\begin{equation}
T^2 = \sum\nolimits_j c_j, 
\quad \text{where} \quad 
c_j = (\ve{x} - \ve{\bar{x}})_j \cdot 
\left[ \ma{S}^{-1} (\ve{x} - \ve{\bar{x}}) \right]_j.
\label{eq:contrib}
\end{equation}

However, the individual contributions $c_j$ may be negative due to the off-diagonal elements of $\ma{S}^{-1}$, which complicates their physical interpretation.

Therefore, for data analysis we adopted a simplified but transparent alternative. First, similar to \cite{Mazzoni2024}, we perform a $z$-normalization for all sensors and temporal clusters, i.e.,
\begin{equation}
z_{jt} = \frac{x_{jt} - \bar{x}_{j,\tau(t)}}{\sigma_{j,\tau(t)}},
\label{eq:znorm}
\end{equation}
which puts all sensors on a common scale but ignores correlations between them (as in weighted least squares). 

Then, the squares of these normalized values $z^2_{jt}$ are propagated to the entire network by Laplacian interpolation. Specifically, letting  $S \subset V$ the set of sensor nodes with given values $f(s)$, and $N(v)$ the set of neighbors of $v$, we solve for unknown nodes $v \in V \setminus S$ the discrete Laplace equation:
\begin{equation} 
    f(v) = \frac{1}{|N(v)|} \sum_{u \in N(v)} f(u),
    \label{eq:laplace}
\end{equation}
with fixed $f(s)$ for $s \in S$. 

Examples of such pre-localization are given in Section~\ref{sec:classify}.

\section{Case Study}
\label{sec:data}
To assess the performance of the suggested methodology, we use the L-Town benchmark from the BattLeDIM 2020 competition~\citep{Vrachimis2022}. 
\subsection{General Description}
\begin{figure} 
    \centering
    \includegraphics[width=1.0\linewidth]{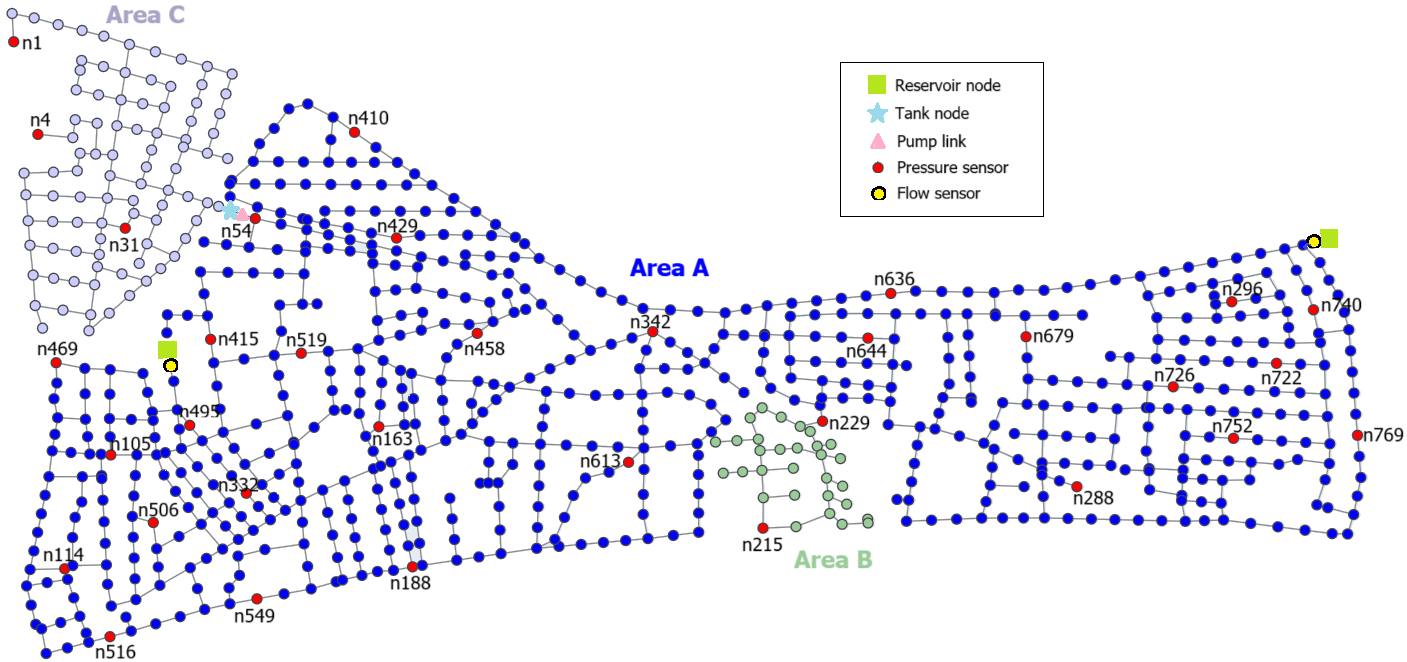}
    \caption{Graphical Representation of the BattLeDIM 2020 L-Town WDN} 
    \label{fig:ltown}
\end{figure}

The L-Town WDN is supplied by two reservoirs and provides drinking water to approximately 10,000 consumers. The network consists of 782 nodes connected by 905 pipes with a total length of 42.6 km. A graphical representation of the network topology is shown in Fig. \ref{fig:ltown}.

The network is partitioned into three district metered areas (DMAs). The largest of them is residential area A, consisting of $\approx 660$ nodes. Area A is connected via a single pipe to two other areas, B and C. 

Commercial area B is located in the lower part of the city and contains around 30 nodes. A pressure reducing valve (PRV) is located at the inlet of the area to maintain a constant pressure. This makes the hydraulic behavior of area B practically independent of other parts of the network. Additional PRVs are located downstream of the two main reservoirs.

The upper part of the city forms the industrial area C with about 90 nodes. This area also has the highest elevation. It is supplied by a water tank equipped with a pump. The pump is programmed so that the tank fills during the night and discharges to area C during the day. Essentially, areas A and C impact each other only through pump operation.

The state of the network is monitored by 33 pressure sensors, most of which (29) are installed in area A. Additionally, three flow meters are installed at the outlets of the two reservoirs and at the pump near the water tank. Area C is equipped with 82 AMRs. Despite such a dense sensor coverage, detection of leaks in area C is not straightforward, as it exhibits the largest by far demand volatility.  

\subsection{Datasets}
Datasets for the BattLeDIM competition simulate two years of network operations and provide a simplified nominal hydraulic model of L-Town in EPANET format. Each network node is characterized by a certain mixture of 
residential, commercial, and industrial consumers.
Demands for each consumer group follow a weekly pattern, 
specified with a 5-minute granularity. 

To convey discrepancies between the real system and its hydraulic model, data for the competition were generated using a modified ``real'' EPANET model, which differed from the nominal model in several important aspects (demand patterns, pipe parameters, etc.). 
This model was complemented with a set of leakage scenarios. Together, they were employed to produce SCADA datasets for 2018 and 2019,
containing data from all available sensors collected with a 5-minute frequency.

The dataset for 2018 also included information for 14 leakages, 10 of which were found and fixed within this year. These data were intended for model training purposes and included pipe ID, start and repair times, and leakage outflows in m$^3/$h. 
An additional 19 leakages started in 2019, summing up to a total of 23 unknown leaks that were to be identified and localized by competition participants.

The simulated leaks were of two types, abrupt and incipient. Abrupt leaks may correspond to pipe bursts that occur suddenly and reach their maximum magnitude nearly instantly. Incipient leaks evolve over time, gradually reaching and stabilizing at their maximum magnitude if undetected. 

The characteristics of individual leakage events are summarized in Table~\ref{tab:leak_events} in the \ref{sec:app}. 
Details on simulating leak dynamics, as well as on other technical issues, can be found in \cite{Vrachimis2022}.
We employ the 2018 datasets for training purposes, while information on 2019 leakages is only used for validation.

\subsection{Descriptive Statistics}
\begin{figure}[ht] 
  \centering
  \begin{subfigure}[c]{0.48\textwidth}
    \includegraphics[width=\textwidth]{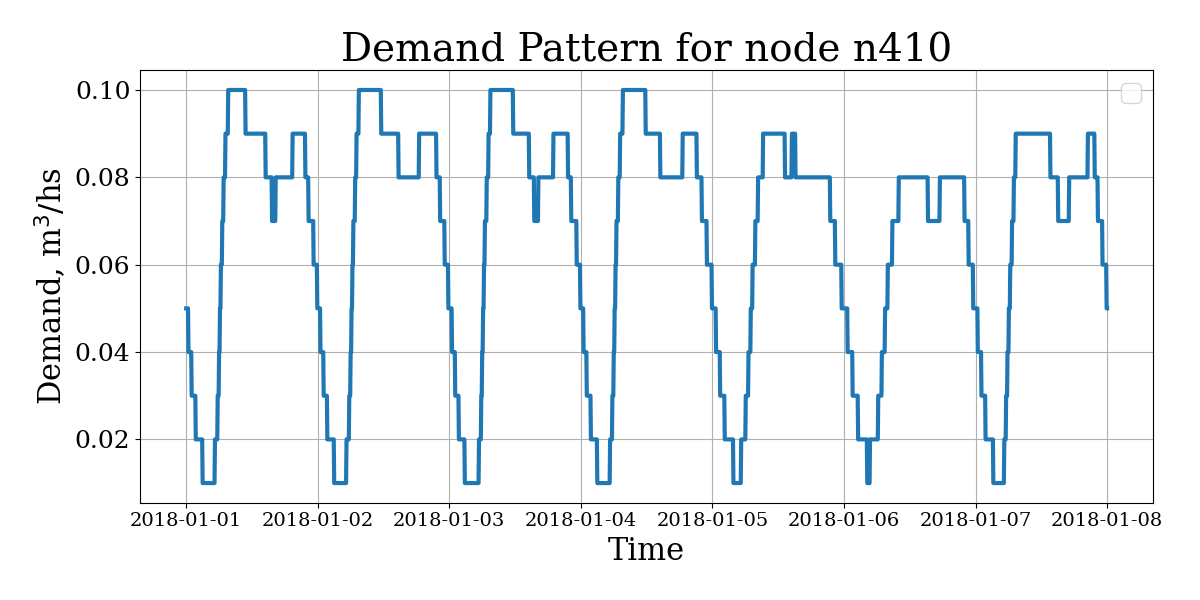}
    \caption{}
    \label{fig:demand410}
  \end{subfigure}
  \hfill
  \begin{subfigure}[c]{0.48\textwidth}
    \includegraphics[width=\textwidth]{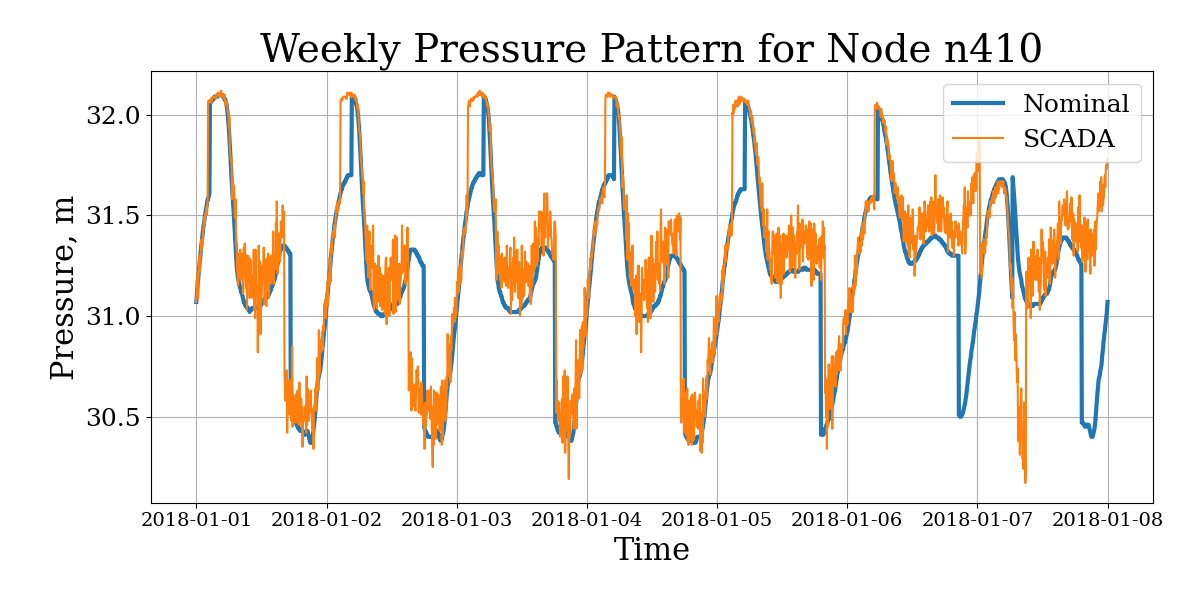}
    \caption{}
    \label{fig:pressure410}
  \end{subfigure}
  \caption{Weekly patterns of demand (a) and pressure (b) for node n410. Blue lines show values from nominal EPANET model, orange line -- SCADA data.}
  \label{fig:n410}
\end{figure}

Demand for water follows a periodic weekly pattern (see Fig.~\ref{fig:demand410} for a typical example). Workdays and night hours are similar, while the consumption pattern over the weekend is somewhat different. There is also a relatively weak seasonality effect, with water consumption increasing in summer and decreasing in winter.

Because higher consumption leads to a loss of head in the network, nodal heads and pressures also exhibit predictable periodic oscillations, shown in Fig.~\ref{fig:pressure410}. The latter figure also illustrates the magnitude of differences between the nominal model and SCADA data.     

\begin{figure}[h] 
  \centering
  \begin{subfigure}[c]{0.65\textwidth}
    \includegraphics[width=\textwidth]{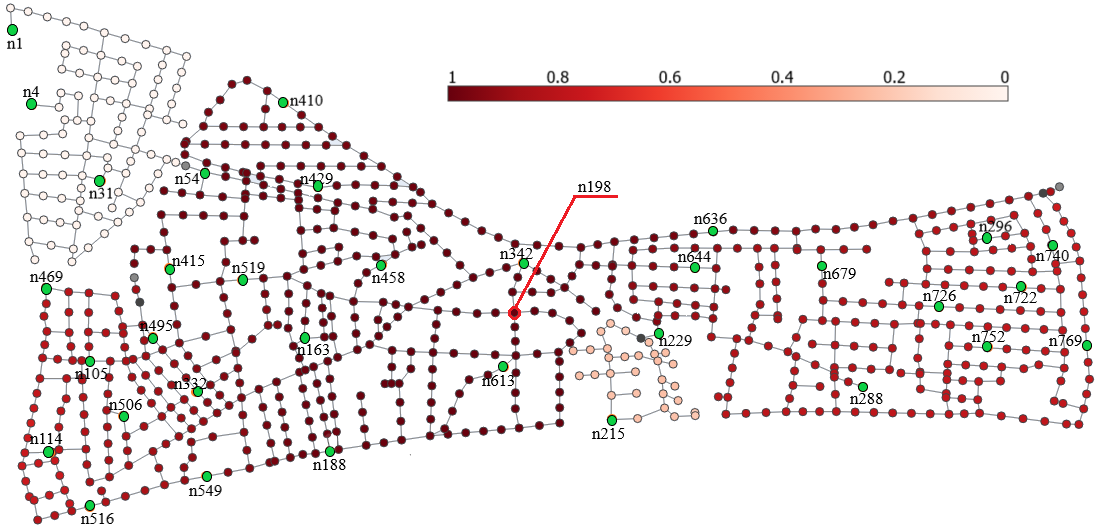}
    \caption{}
    \label{fig:corrmap}
  \end{subfigure}
  \hfill
  \begin{subfigure}[c]{0.33\textwidth}
    \includegraphics[width=\textwidth]{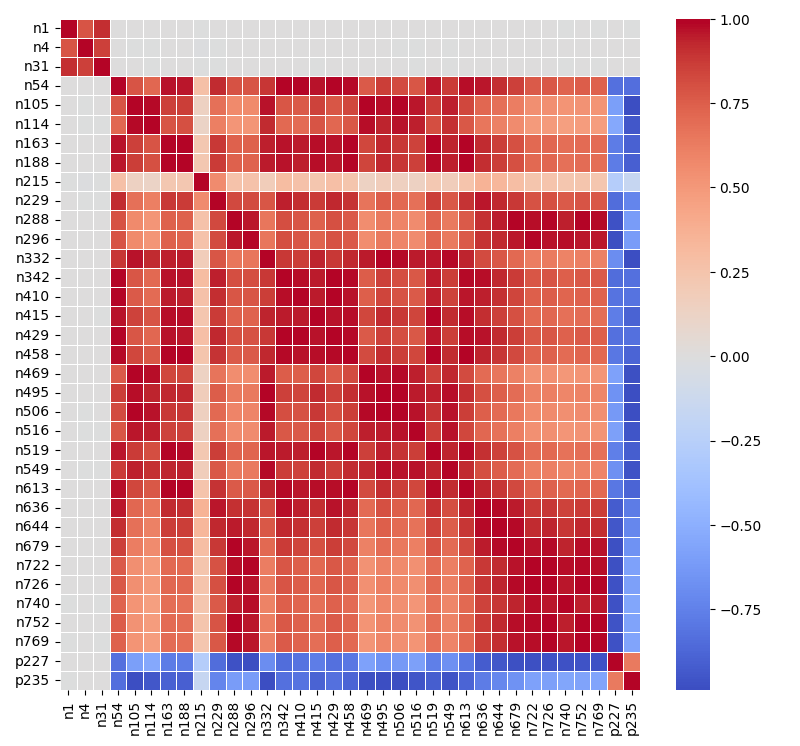}
    \caption{}
    \label{fig:heatmap}
  \end{subfigure}
  \caption{Graphical representation of the network spatial correlation structure. 
  }
  \label{fig:corr}
\end{figure}

Another distinctive feature of the data is the strong spatial correlation between the monitored parameters (pressure, water flow rate), illustrated in Fig.~\ref{fig:corr}. Fig.~\ref{fig:corrmap} shows the value of the correlation coefficient between the centrally located node n198 and all the others by gradient coloring. Fig.~\ref{fig:heatmap} shows the correlation matrix for readings of 33 pressure sensors and two flow sensors in the form of a heatmap. Both figures correspond to the nominal L-Town model evaluated at 2018-01-01 00:00 and were constructed by adding a multiplicative 10\% Gaussian noise to nodal demands.

Both of these figures confirm the observations made earlier: there is a very weak correlation between the pressure values in area A, area B (pressure sensor n215), and area C (pressure sensors n1, n4, and n31). However, there is a very strong correlation between pressures within each of these areas. Correlations between pressures and flow rates are negative, as can be expected based on physics considerations. 

\begin{figure}[h]
  \centering
  \begin{subfigure}[c]{0.48\textwidth}
    \includegraphics[width=\textwidth]{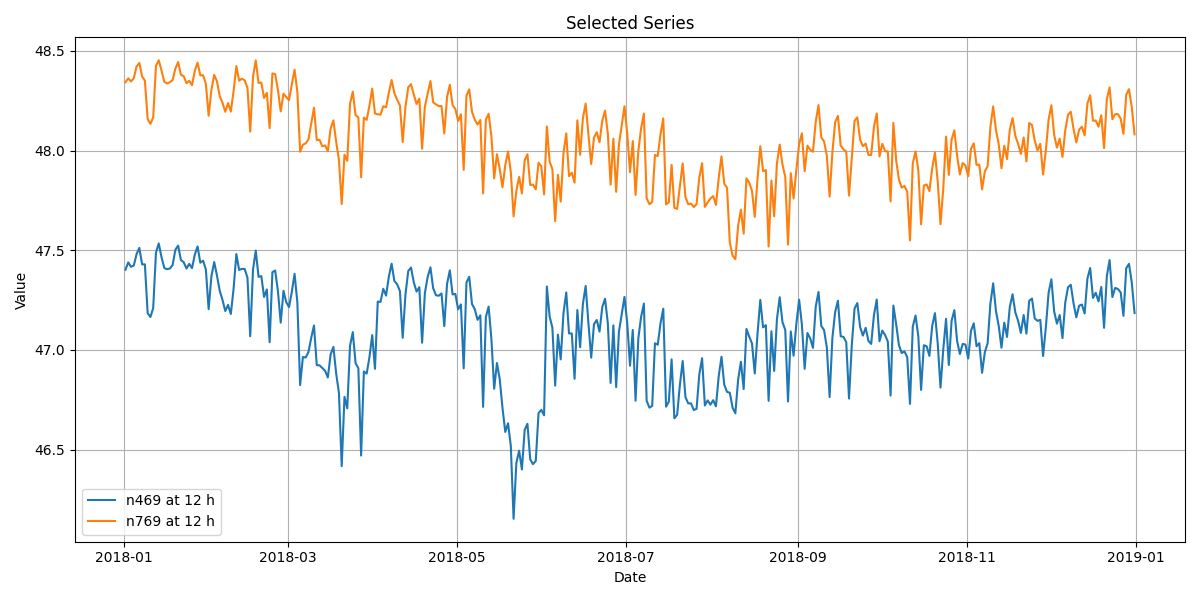}
    \caption{}
    \label{fig:cs2018}
  \end{subfigure}
  \hfill
  \begin{subfigure}[c]{0.48\textwidth}
    \includegraphics[width=\textwidth]{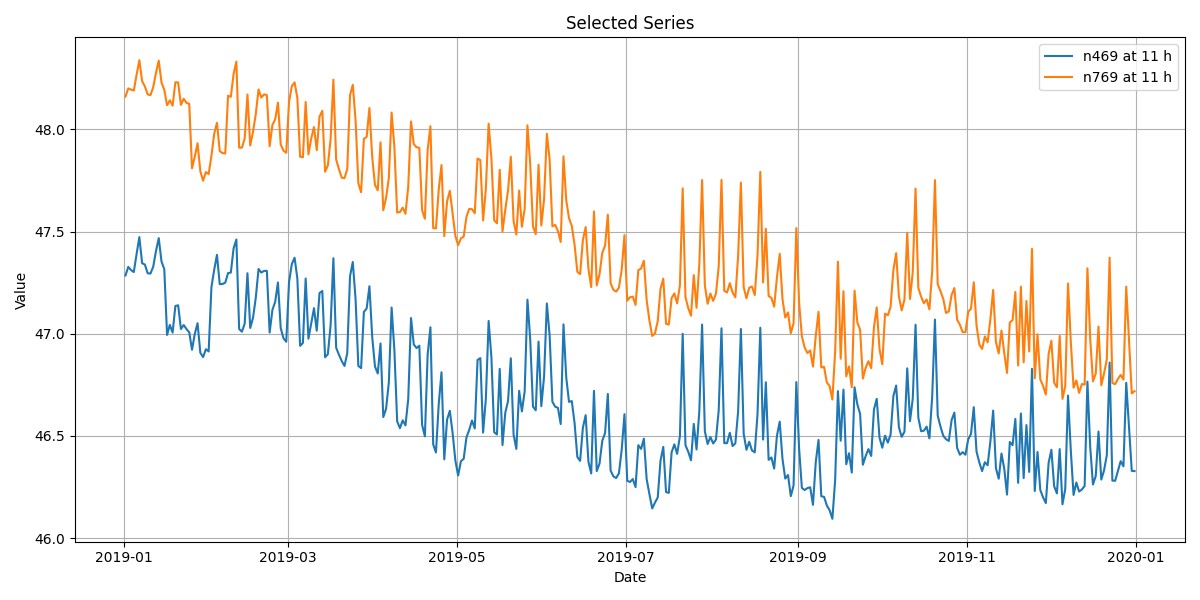}
    \caption{}
    \label{fig:cs2019}
  \end{subfigure}
  \caption{Time series of pressure values for sensors at nodes n469 and n769 resampled at hourly frequency at 11:00 am in 2018 (a) and 2019 (b) SCADA datasets.}
  \label{fig:cs}
\end{figure}

High correlation between the sensor readings within the certain area is also evident from the SCADA dataset. Fig.~\ref{fig:cs2018} shows 2018 time series of meter readings for two pressure sensors, n469 and n769, resampled at hourly frequency at 11:00 am. Despite the fact that these two sensors are located at the opposite ends of the map, the similarity of patterns between them is quite strong and becomes even stronger in 2019 dataset (Fig.~\ref{fig:cs2019}).  

High degree of correlation between readings of geographically distant sensors complicates detection and especially localization of anomalies. Our proposed approach eliminates these correlations by using Mahalanobis metrics.

\section{Results and Discussion}
\label{sec:results}

The proposed framework was validated using the L-Town network as a test environment, combining both simulation-based experiments and tests on SCADA data. Certain aspects of performance evaluation, such as the the agreement between the empirical behavior of the test statistic and its theoretical distributional properties, or the assessment of localization accuracy, can only be examined through simulation under controlled conditions. Therefore, the first stage of validation consisted of computer simulations in WNTR/EPANET with synthetic noise and systematically introduced leaks.

Subsequently, the framework was applied to the SCADA dataset provided in the BattLeDIM competition. The resulting leak detection and localization outputs were compared against the competition benchmarks and with results reported in later published studies. 

\subsection{Accounting for Sensor Correlations in Anomaly Detection}
\label{sec:calibration}
Anomaly detection in WDN is typically performed using data from multiple pressure and flow sensors rather than from a single measurement point. This multivariate nature must be taken into account during the calibration of the detection system.

If sensor readings were statistically independent, the overall false alarm rate could be easily calculated from the univariate probabilities. For instance, denoting the individual false alarm probability for each sensor by $p$, the overall probability of at least one false alarm among $s$ sensors would be:
\begin{equation}
    p_\text{FA}=1-(1-p)^s, \text{ from where } p=1 - \sqrt[s]{1 - p_\text{FA} 
}.
    \label{eq:false_alarm}
\end{equation}

However, in practice, sensor signals in WDN are often highly correlated due to the hydraulic interdependence of the network (see Fig.~\ref{fig:corr}). Since the independence assumption is violated, the true false alarm probability cannot be directly obtained from the univariate rates.

When anomaly detection is performed for each sensor separately (e.g., by flagging any observation that exceeds the 95\% confidence limits of its historical range), the system-level false alarm rate can be significantly higher than the nominal 5\%. There exist several approaches to combat this problem. The simplest and most widely used method is the Bonferroni correction, which suggests setting $p=p_\text{FA,desired}/s$ \citep{Clements2024}. This procedure guarantees that $p_\text{FA,real} \le p_\text{FA,desired}$. However, in the presence of strong correlations among sensor readings, the Bonferroni correction tends to be overly conservative.
As a result, the actual false alarm probability becomes much lower than the nominal level, which can significantly reduce the sensitivity of anomaly detection.

In contrast, Hotelling's $T^2$ statistic \eqref{eq:teststat} accounts for the covariance structure among the sensors, which allows setting the detection threshold to achieve a desired overall false alarm rate.

\begin{figure}[h]
  \centering
  \begin{subfigure}[c]{0.48\textwidth}
    \includegraphics[width=\textwidth]{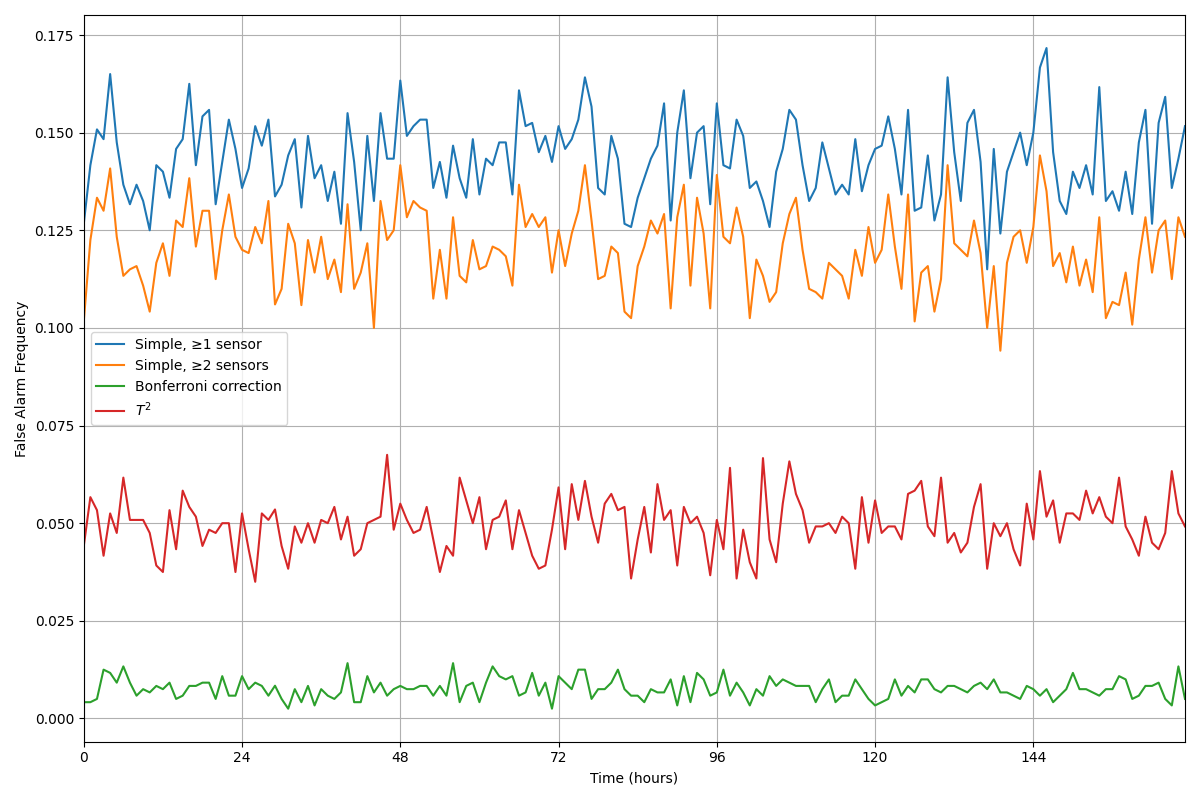}
    \caption{}
    \label{fig:FA_rate}
  \end{subfigure}
  \hfill
  \begin{subfigure}[c]{0.48\textwidth}
    \includegraphics[width=\textwidth]{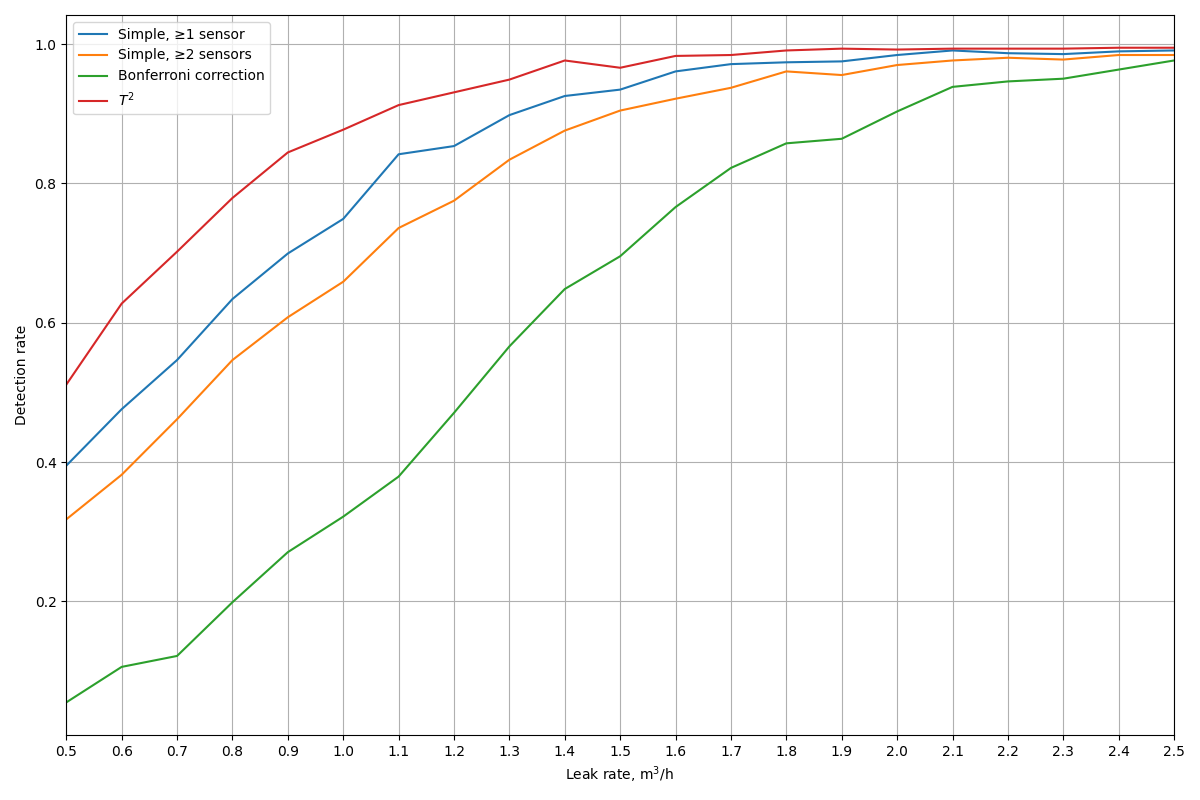}
    \caption{}
    \label{fig:det_rate}
  \end{subfigure}
  \caption{False alarm rate (a) and detection rate (b) for various detection methods.}
  \label{fig:err_rates}
\end{figure}

To experimentally verify these considerations, a simulation of the L-Town WDN was conducted for one week using WNTR/EPANET. To model uncertainty, a multiplicative Gaussian noise with a standard deviation of 10\% was applied to the nodal demands.
Fig.~\ref{fig:FA_rate}  presents false alarm frequencies for 100 simulations  
obtained using several anomaly detection strategies:

(1) detection when at least one sensor exhibits anomalous readings;

(2) detection based on at least two sensors;

(3) detection with the Bonferroni correction; and

(4) detection using the Hotelling's $T^2$ statistic.

It can be observed that the overall false alarm rate for the first two methods is considerably higher than the desired 5\% level, while the Bonferroni correction yields a rate that is substantially lower. In contrast, the Hotelling's $T^2$ approach achieves a false alarm probability that closely matches the target significance level.

Fig.~\ref{fig:det_rate} presents the leak detection rates for several detection methods across a range of leak magnitudes. In the experiment, leaks of varying magnitudes were simulated in all pipes within Area A under 10\% Gaussian noise. Based on these simulations, the leak detection probability was evaluated as a function of the leak magnitude. Detection was performed using the hydraulic state of the network at 00:00 on Monday. The results show that the Hotelling-based detector outperforms alternative approaches, enabling reliable detection of considerably smaller leaks.

\subsection{Localization of leakages}
\label{sec:localization}
To validate the leak localization algorithm, the following experiment was conducted. Multiplicative Gaussian noise with a predefined standard deviation was added to the nodal demands. Then, a leak of a given magnitude was sequentially simulated in every pipe within the study area. The geometric distance between the simulated leak and the point of maximum value on the interpolated field (corresponding to the most probable leak location) was computed. If this distance did not exceed 300 meters, the leak was considered successfully localized, following the BattLeDIM competition rules. The results of the experiment are summarized in Table \ref{tab:combined_localization}: (a) average geometric localization error; and (b) percentage of successful detections.

\begin{table}[h!]
\centering
\caption{Localization performance under different noise levels and leak sizes}

\begin{minipage}{0.48\linewidth}
\centering
\caption*{(a) Mean geometric localization error, m}
\begin{tabular}{c|ccc}
\toprule
Leak size, & \multicolumn{3}{c}{Noise level} \\
m$^3$/h & 5\% & 10\% & 20\% \\[-0.5em]
\midrule
\ 2  & 153 & 165 & 340 \\
\ 5  & 150 & 153 & 154 \\
10 & 150 & 150 & 152 \\
\bottomrule
\end{tabular}
\end{minipage}
\hfill
\begin{minipage}{0.48\linewidth}
\centering
\caption*{(b) Successful localization rate, \%}
\begin{tabular}{c|ccc}
\toprule
Leak size, & 
\multicolumn{3}{c}{Noise level} \\
m$^3$/h & 5\% & 10\% & 20\% \\[-0.5em] 
\midrule
\ 2  & 91.4 & 90.9 & 74.9 \\
\ 5  & 92.1 & 92.0 & 91.7 \\
10 & 92.0 & 92.1 & 91.7 \\
\bottomrule
\end{tabular}
\end{minipage}
\label{tab:combined_localization}
\end{table}


It should be noted that, according to the maximum principle for harmonic functions, Laplacian interpolation cannot produce local extrema within the domain. The maximum of the interpolated field always occurs at one of the support nodes, i.e., the sensor locations. Therefore, the described procedure is effectively equivalent to verifying whether the most out-of-range sensor lies within a 300-meter radius of the leak position.

As expected, higher noise levels make localization more difficult, while increasing the leak magnitude improves it. However, due to the nature of the method, the localization accuracy does not further improve once the leak becomes sufficiently pronounced. Likewise, the proportion of leaks that are successfully localized settles at approximately 92\%. The other $\approx$8\% of instances could be due to the difference between geometric distance and actual network (hydraulic) distance, potentially impacting sensor response sensitivity in a non-linear manner.

The experiment was conducted only for Area A. For Areas B and C, due to their small size, detecting a leak within the area is practically equivalent to localizing it according to the BattLeDIM criteria.

\begin{figure}[ht] 
  \centering
  \begin{subfigure}[c]{0.49\textwidth}
    \includegraphics[width=\textwidth]{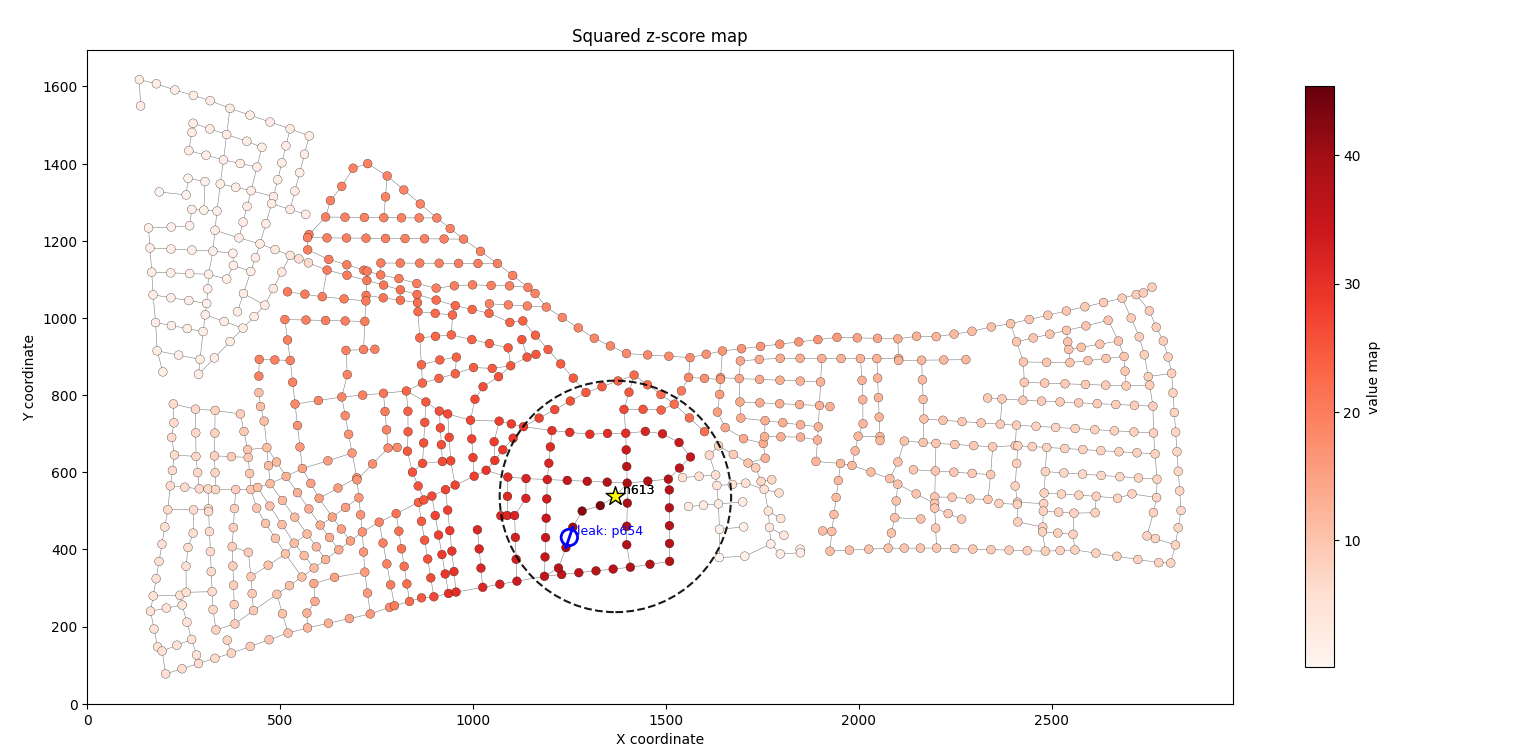}
    \caption{}
    \label{fig:1leak}
  \end{subfigure}
  \hfill
  \begin{subfigure}[c]{0.49\textwidth}
    \includegraphics[width=\textwidth]{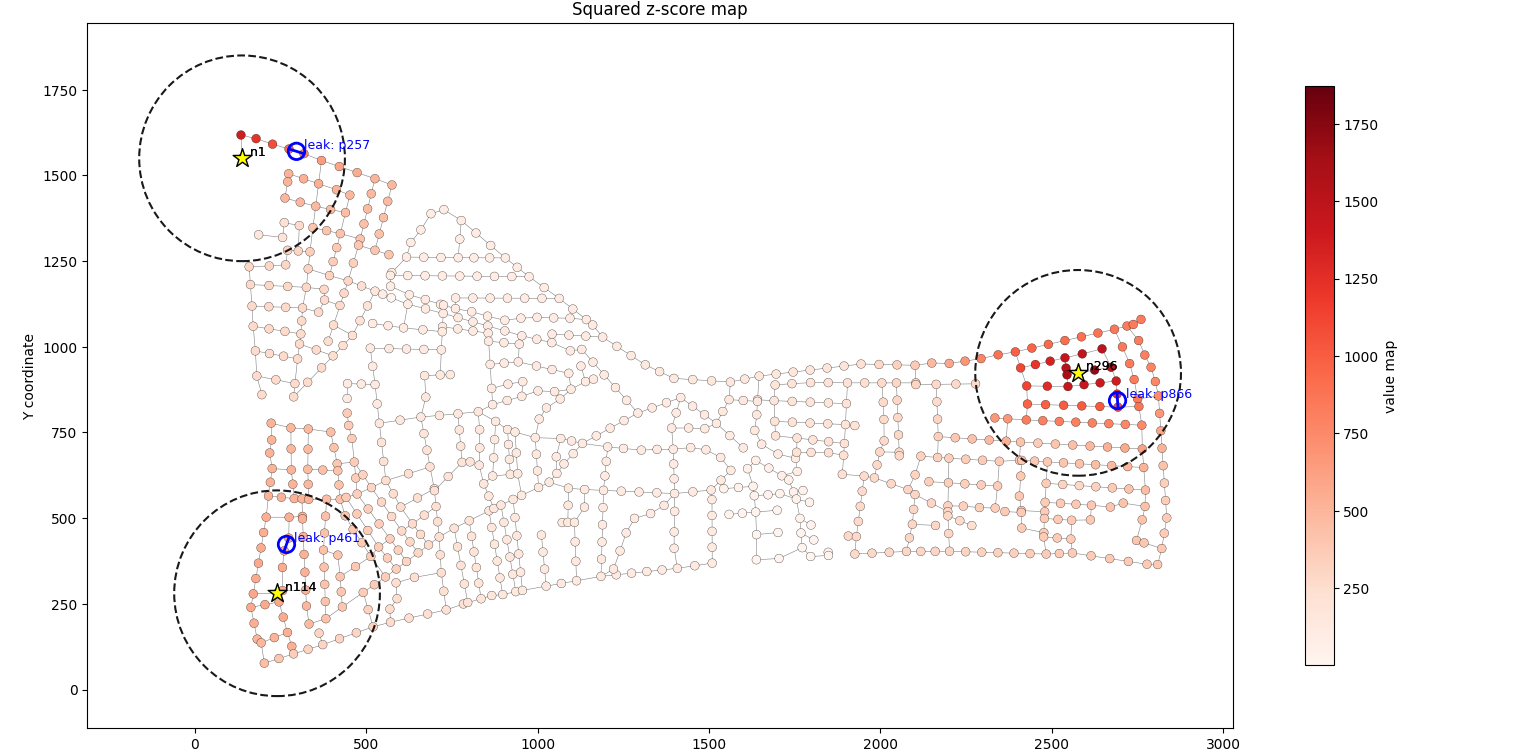}
    \caption{}
    \label{fig:3leaks}
  \end{subfigure}
    \begin{subfigure}[c]{0.49\textwidth}
    \includegraphics[width=\textwidth]{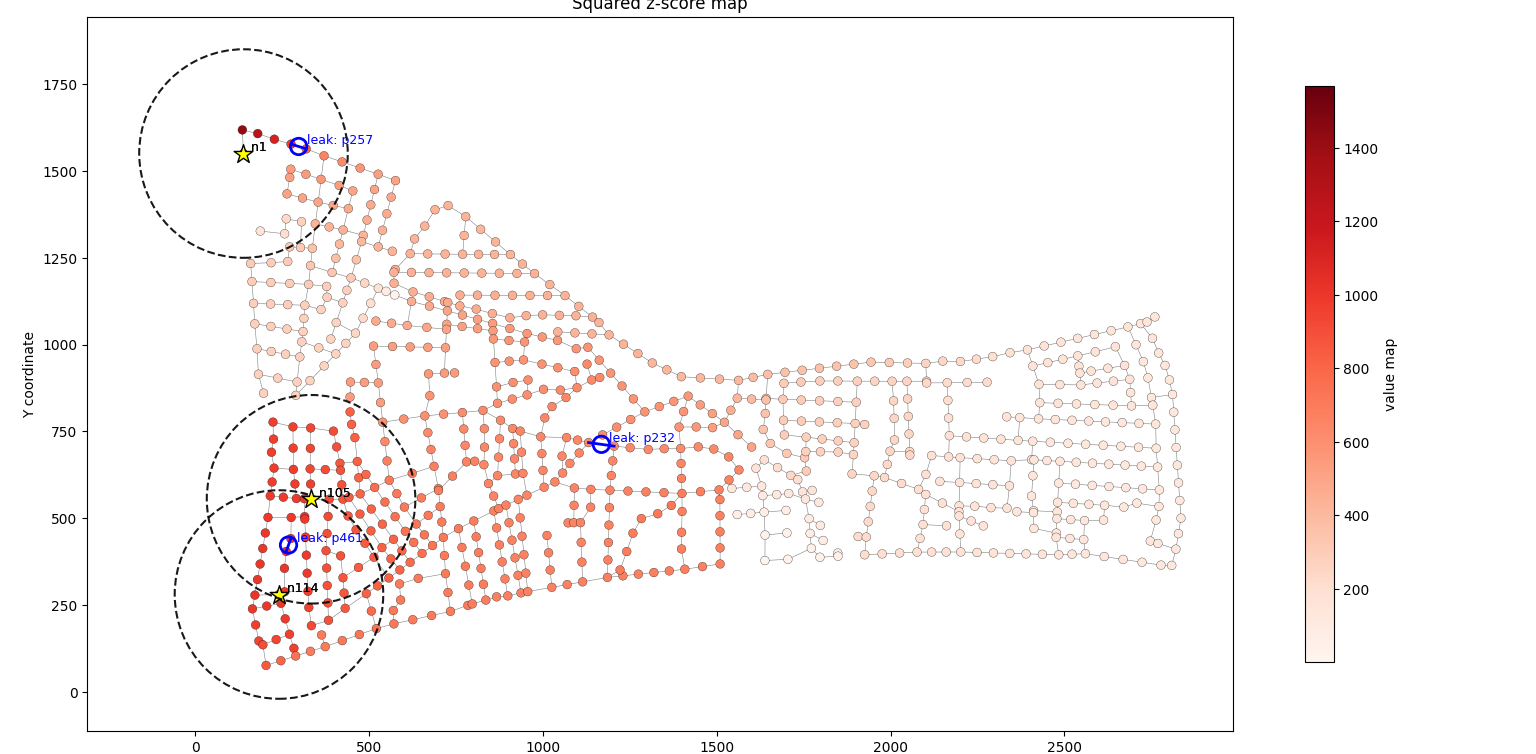}
    \caption{}
    \label{fig:3leaks_fail}
  \end{subfigure}
  \hfill
  \begin{subfigure}[c]{0.49\textwidth}
    \includegraphics[width=\textwidth]{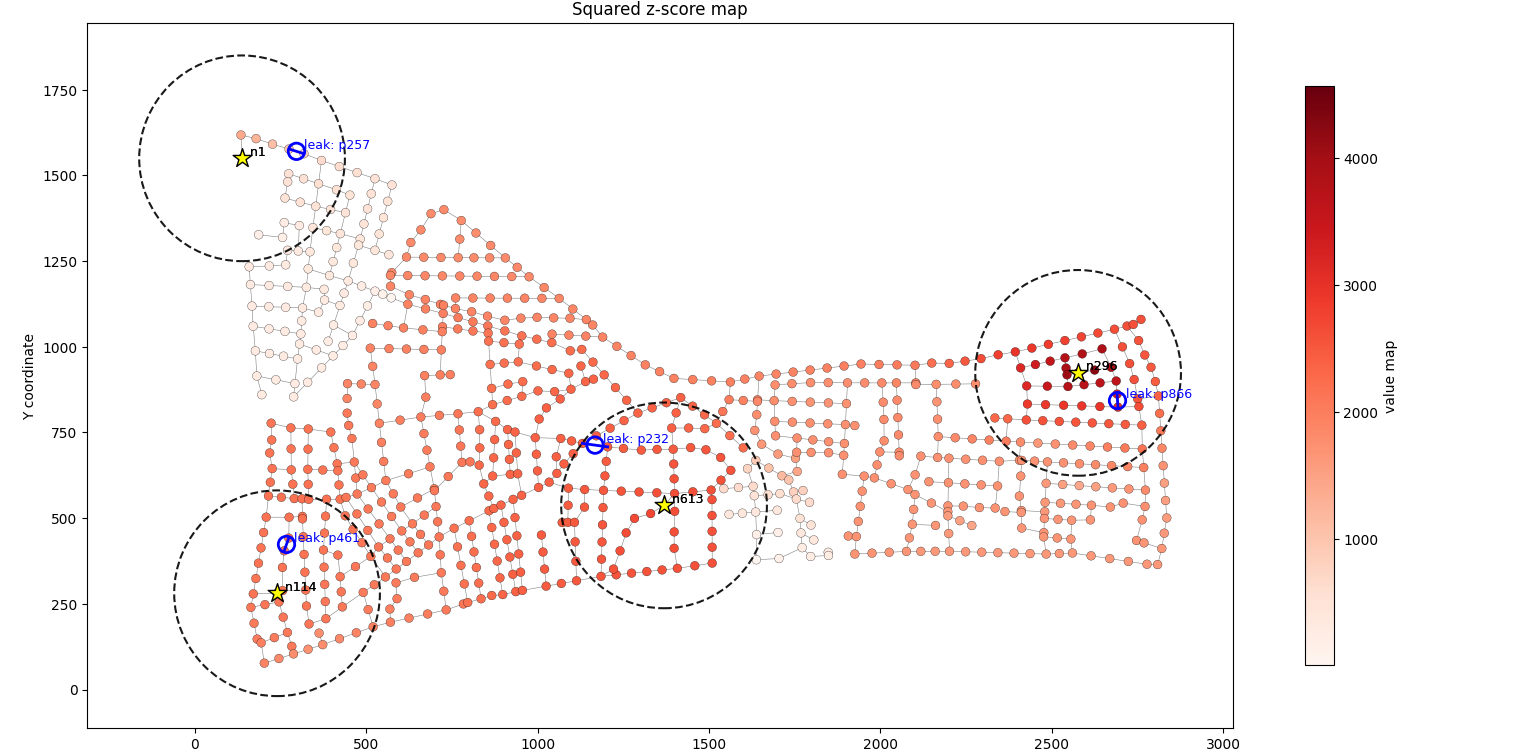}
    \caption{}
    \label{fig:4leaks}
  \end{subfigure}
  \caption{Pre-localization of selected leakage events from 2018 data set. Asterisks mark nodes with the maximum $z^2$ value in the interpolated field after iterative suppression of previously detected anomalies. Small circles indicate true leak locations. Dashed circles correspond to the 300\,m error margin considered acceptable by the BattLeDIM rules.}
  \label{fig:leakloc}
\end{figure}

An example of the successful localization of a single leak is provided in Fig.~\ref{fig:1leak}, where the scalar field indicating the likelihood of leak occurrence in different network nodes is visualized on the network map using a color gradient. The figure also shows the true leak location, the estimated leak position, and the 300-meter tolerance radius around the latter, in accordance with the BattLeDIM evaluation criteria.

In the case of multiple simultaneous leaks, the localization task becomes more challenging. When the leaks are spatially well-separated, the Laplacian interpolation typically produces a field map containing several distinct “hot spots.” These regions can be identified either visually or algorithmically, e.g., by iteratively selecting the sensor with the highest deviation and then excluding the corresponding neighborhood from subsequent searches. An example of successful localization using this technique is shown in Fig. \ref{fig:3leaks}.

However, when leaks are located close to one another or differ significantly in size, the smaller leak may be overshadowed by the hydraulic effects of the larger one. A representative example is the incipient leak p232. For small discharge rates, it produces only marginal deviations in the neighboring sensors, which remain dominated by the anomalies caused by other, larger leaks, and the leak cannot be correctly localized (Fig.~\ref{fig:3leaks_fail}). As the leak magnitude increases, its contribution becomes sufficiently pronounced to be distinguishable even in the presence of three other leaks (Fig.~\ref{fig:4leaks}).

Overall, while iterative suppression of known leaks can enhance multi-leak localization in favorable conditions, the reliability of this approach remains limited. More stable performance can be achieved by re-training the anomaly detection model on updated data, effectively incorporating detected leaks into a new baseline operating state. Following this, anomaly detection is carried out relative to the revised reference, allowing for the recognition of incremental changes to the new status quo.

Extending this method to more accurately account for individual sensor contributions, e.g., using the Shapley value \citep{ghorbani19}, may be a promising direction for future work.

\subsection{Discrimination between Leaks and Faulty Sensors}
\label{sec:classify}
As an illustrative case for anomaly classification, we consider the problem of distinguishing between leaks and faulty sensors. To develop a discriminative rule, 
a simulation study was conducted in WNTR by introducing leaks of varying sizes at different nodes within area A of the L-Town benchmark. 
To emulate realistic operating conditions, a multiplicative Gaussian demand noise of 10\% was applied to all nodes. For each simulated scenario, the number of sensors exceeding the admissible operational range, $M_a$, was recorded.

\begin{figure}[t]
    \centering
    \includegraphics[width=1.0\linewidth]{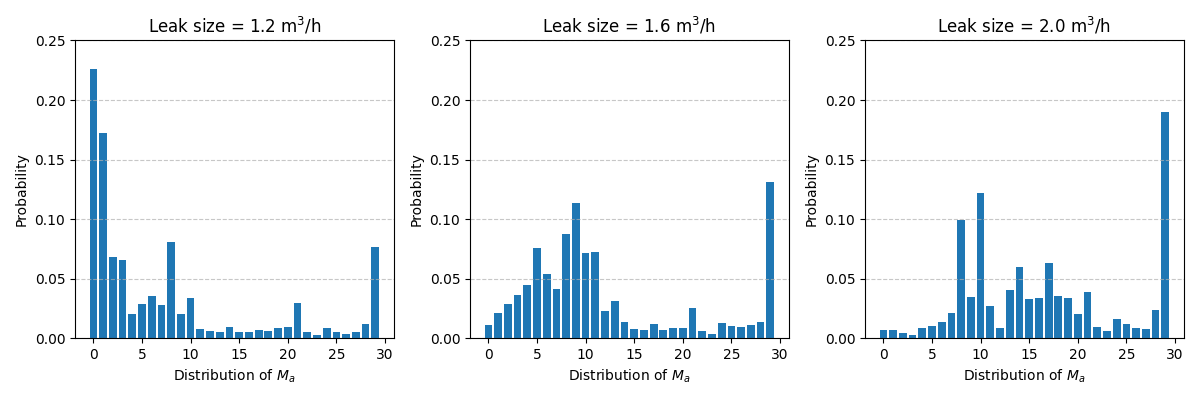}
    \caption{Distribution of number of out-of-range sensor count $M_a$ at various leak sizes (only pressure sensors were considered in the simulation experiment)}
    \label{fig:countdist}
\end{figure}

Fig.~\ref{fig:countdist} presents the histogram of $M_a$ distribution for different leak sizes. The results show that the number of out-of-range sensors increases sharply with leak magnitude. At 1.2 m$^3$/h, the most likely outcome is that not a single sensor would exceed its operational range. Leaks of 1.6 m$^3$/h or greater almost invariably cause out-of-range readings in multiple sensors, while leaks of greater magnitude are likely to produce anomalous readings across the entire area. Therefore, if Hotelling's $T^2$ statistic indicates an anomaly but only a single sensor deviates, a systematic sensor error is likely. For smaller leaks, which are unlikely to indicate a pipe burst, deviations in additional sensors are expected over time; the absence of such corroborating evidence suggests a sensor-related anomaly.

An alternative approach for distinguishing a leak from a biased sensor is based on the same idea as the pre-localization method described above. For each sensor $j$, we compute the normalized deviation $z_j = (x_j - \bar{x}_j)/\sigma_j$, and evaluate the overall anomaly magnitude as $Z^2 = \sum_{j} z_j^2$.
To characterize the spatial distribution of anomalous behavior, we quantify the relative contribution of each sensor as
$s_j = z_j^2/Z^2$.

In scenarios involving a systematic measurement bias, the distribution of $s_j$ tends to be highly concentrated at the malfunctioning sensor. Conversely, in the presence of a hydraulic leak, the contributions $s_j$ exhibit a spatially decaying pattern with increasing distance from the leak location. 

\begin{figure}[!h]
    \centering
    \includegraphics[width=0.6\linewidth]{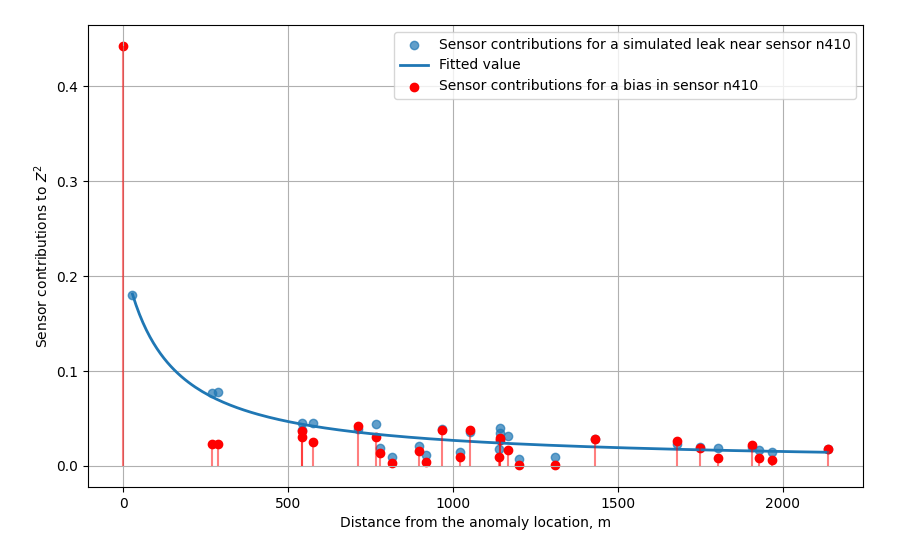}
    \caption{Comparison of sensor contribution profiles $s_j$ under leak-induced and bias-induced anomalies.}
    \label{fig:contributions}
\end{figure}

Fig.~\ref{fig:contributions} demonstrates this behavior by contrasting the distributions of $s_j$ obtained for a 0.02\,m bias applied to sensor n410 and for a 1.9\,m$^3$ leak introduced in the adjacent pipe p350. Under the selected simulation conditions, both cases yield comparable values of the Hotelling's $T^2$ statistic, sufficiently high to suggest an anomaly. However, the spatial signatures of $s_j$ remain clearly distinguishable, enabling reliable classification between sensor bias and leak-induced anomalies.

\subsection{Method Validation using SCADA data}
Next, we examine the application of the proposed leak detection methodology to the SCADA dataset used in the BattLeDIM competition. A summary of the leak events that occurred during the observation period is presented in Table~\ref{tab:leak_events}.

Table~\ref{tab:leak_events} illustrates that selecting a period of nominal operation for training the model in this benchmark is nontrivial, since the first leak occurred as early as the second week and remained unresolved until the final period. We therefore consider model training and validation separately for each of the three DMAs in L-Town.

The simplest region to analyze is area~B, whose state is monitored by a single sensor, n215. This sensor is largely insensitive to changes in the rest of the system and exhibits consistently low variance. Consequently, all time intervals for this sensor were assigned to a single temporal cluster. During 2018, area~B experienced only one leak event, associated with pipe p673. Time intervals outside the leak period were used for model training.

Accordingly, the whitening transformation for area~B reduces to scalar $z$-scaling of the form $z_t = (x_t - \mu)/\sigma$. 

\begin{figure} 
    \centering
    \includegraphics[width=0.95\linewidth]{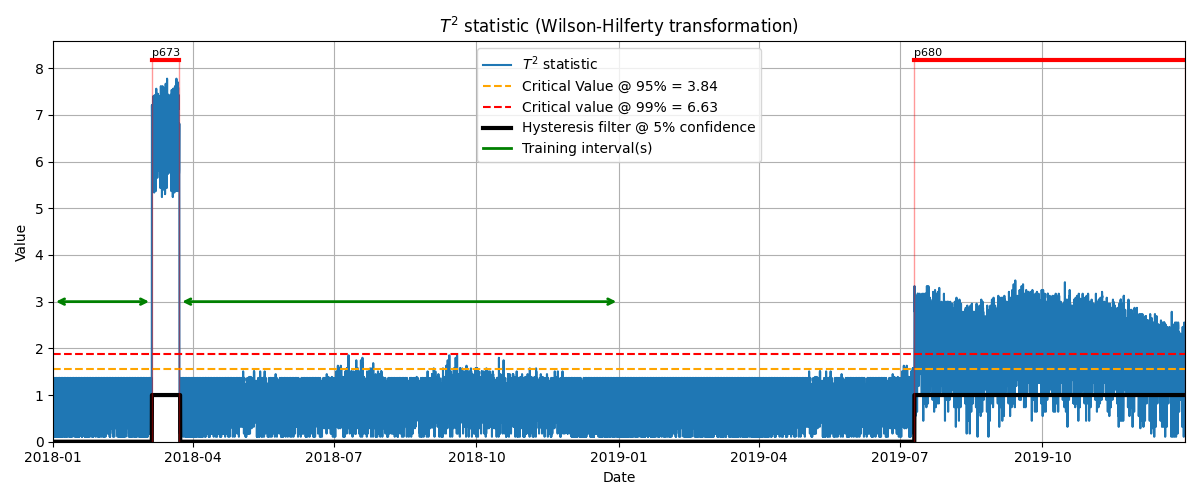}
    \caption{Leak detection in area B for 2018 and 2019 SCADA data}
    \label{fig:areaB}
\end{figure}

The results of applying the methodology to the training (2018) and validation (2019) sets are presented in Fig.~\ref{fig:areaB}. To ensure commensurability, here and below we plot the data using the Wilson–Hilferty cube root transformation $(\chi^2(s)/s)^{1/3}$, which is used to symmetrize and approximately normalize chi‑squared distributed variables \citep{wh1931}. It distorts the shape of the $Z^2$ distribution less than a logarithmic scale and also avoids negative values. 

As the identification relied solely on data from area B, the leak incident is confined to the same area.

Area C differs from other DMAs in that its water supply is regulated by a storage tank (T1) with a variable water level, controlled by a pump. This configuration introduces fluctuations in the hydraulic head, thereby affecting the pressure across the network. Consequently, we modeled the operational state of area C using a four-dimensional feature vector, where three components correspond to pressure sensors (n1, n4, and n31), and the fourth represents the water level in tank T1.

\begin{figure}[h!] 
    \centering
    \begin{subfigure}[c]{0.95\textwidth}
        \includegraphics[width=1.0\linewidth]{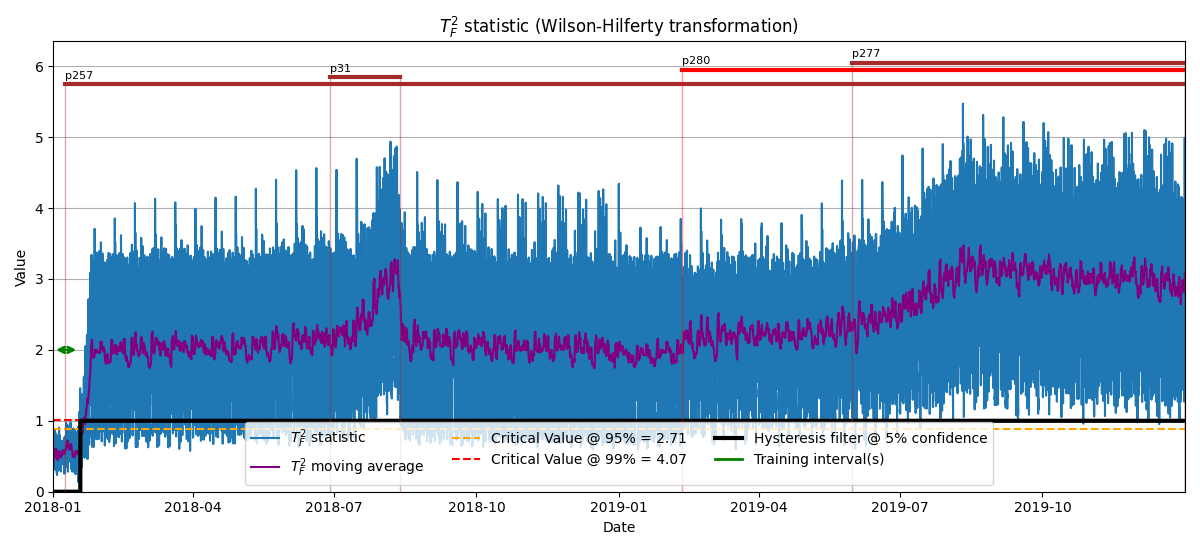}
        \caption{}
        \label{fig:areaC1}
    \end{subfigure}
    \begin{subfigure}[c]{0.95\textwidth}
        \includegraphics[width=1.0\linewidth]{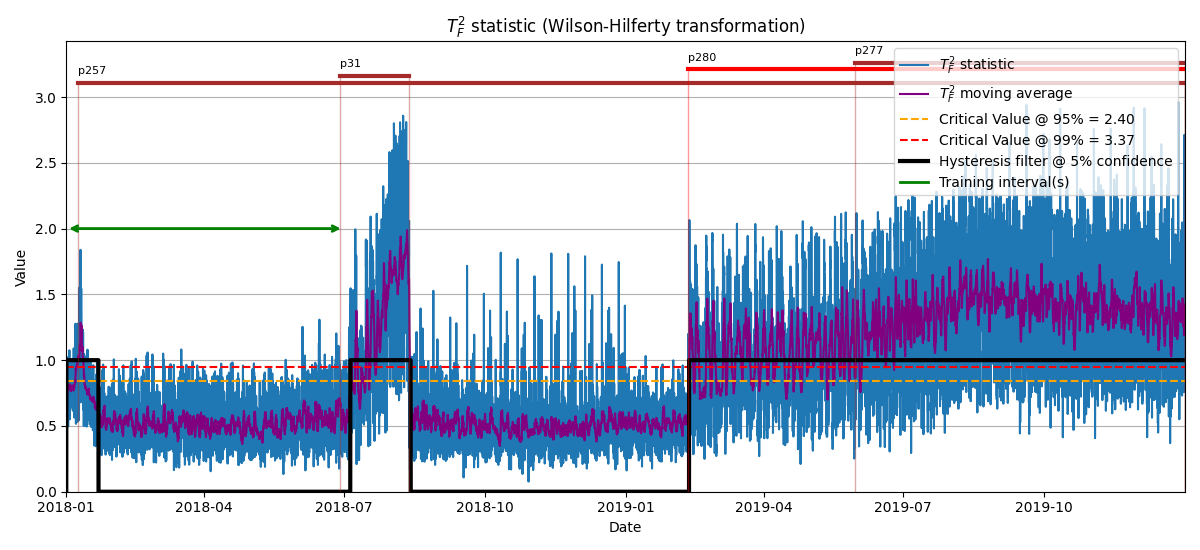}
        \caption{}
        \label{fig:areaC2}
    \end{subfigure}
    \label{fig:areaC}
    \caption{Leak detection in area C for 2018 and 2019 
    using different training periods}
\end{figure}

In 2018, area C experienced two leakage events. The first event (p257) began on January 8 at 13:30 and remained undetected until the end of 2019. As a result, the duration of the normal operating mode is insufficient for collecting representative statistical data. For training purposes, the interval from the beginning of the year to 2018-01-17 18:00 was selected, during which the leakage volume stayed below 2 m$^3$/h. Observations were grouped by two-hour intervals. Due to the limited number of observations, the rescaled Hotelling's statistic $T^2_F$, as defined in Equation~\eqref{eq:t2f}, was employed. Fig.~\ref{fig:areaC1} illustrates the evolution of this statistic over time, along with its 12-hour moving average and the result of applying the hysteresis filter. Known leak events are shown using a Gantt diagram, with abrupt leaks highlighted in red and incipient leaks in brown.

As shown in Fig.~\ref{fig:areaC1}, leak p257 was promptly detected: the filter switched to an anomalous state and remained in that state until the end of 2019. While this response is technically correct, it hinders the detection of subsequent leakages. A retraining approach to identifying these events is illustrated in Fig.~\ref{fig:areaC2}, where the model was retrained on the interval preceding the second leak (p31). In this case, the filter retrospectively classifies the period before stabilization of leak p257 as anomalous and successfully detects subsequent leakages p31 and p280. A similar retraining procedure, based on the interval containing two coexisting background leakages, can be applied to detect the final event (p277). 

Last three leaks can also be identified by change-point detection, but the advantages of using this method are more pronounced for area A. For area C, detection results are given in Table~\ref{tab:leak_events}. 




Area A is by far the largest DMA in L-Town. To describe its operational state, we have used as feature vectors data from 29 pressure sensors, 2 water flow sensors, and 1 water tank level sensor.

\begin{figure}[h] 
    \centering
        \begin{subfigure}[c]{1.0\textwidth}
        \includegraphics[width=0.95\linewidth]{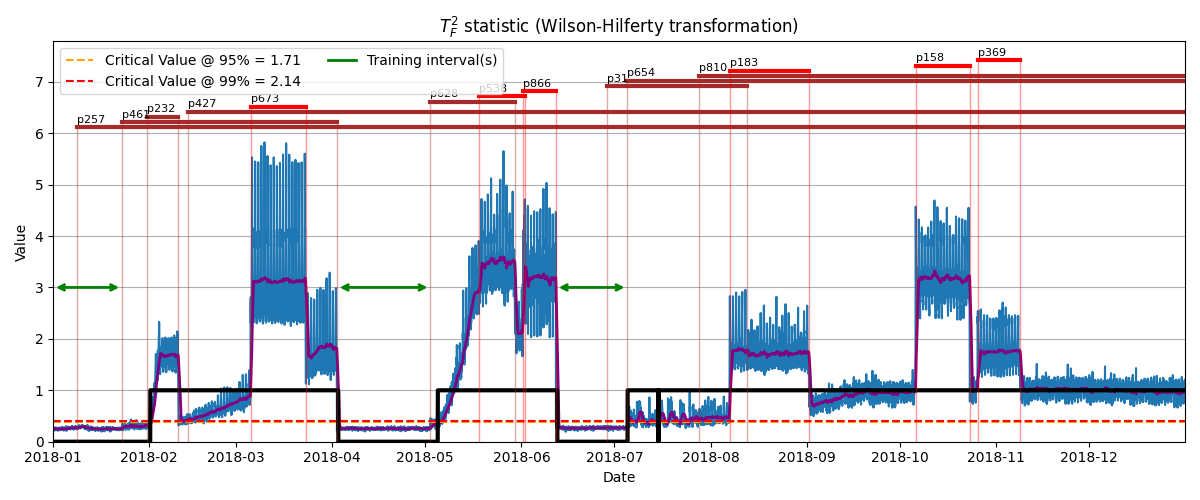}
        \caption{}
        \label{fig:areaA18}
    \end{subfigure}
    \begin{subfigure}[c]{1.0\textwidth}
        \includegraphics[width=0.95\linewidth]{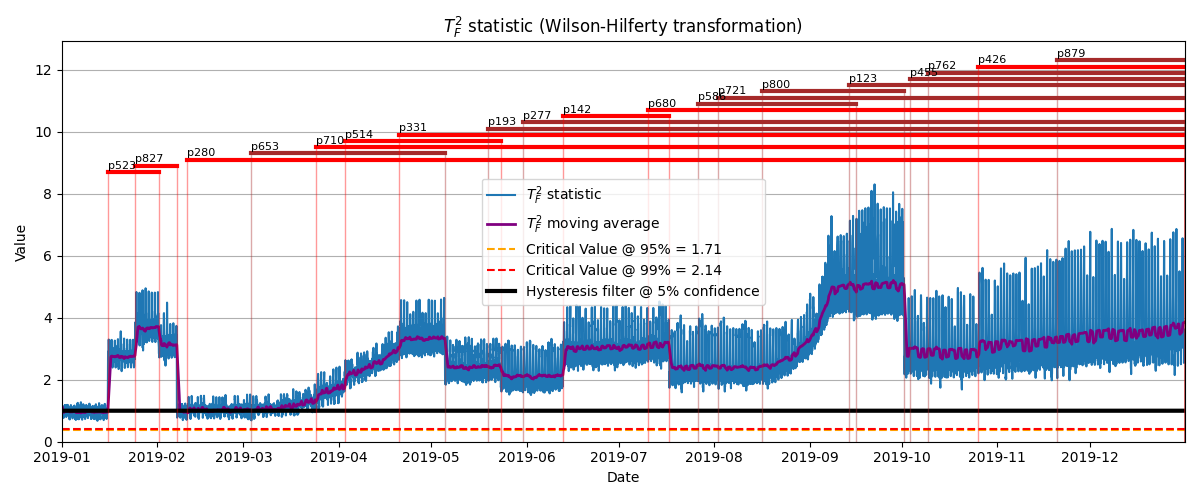}
        \caption{}
        \label{fig:areaA19}
    \end{subfigure}
    \caption{Leak detection in area A for 2018 (a) and 2019 (b) SCADA data}
    \label{fig:areaA} 
\end{figure}

The vast majority of leaks in the L-Town dataset also occurred in area A: 11 in 2018 and 16 in 2019.  It should be noted that while areas B and C can be considered separately from area A, leak events occurring in these areas may affect the hydraulic behavior of area A and cannot be ignored. Thus, as with area C, the limited number of observations under nominal conditions poses a challenge. For model training, we selected periods in 2018 unaffected by leaks, except for the first two undetected cases (p257 and p427). Observations were grouped on an hourly basis. Results of applying the model trained in such a way are shown in Fig.~\ref{fig:areaA}. Data periods used for training are indicated by green arrows.

Due to the large number of leaks, the results for 2018 and 2019 are shown separately (Fig.~\ref{fig:areaA18}, \ref{fig:areaA19}).
The figure 
illustrates how Hotelling's $T^2_F$ statistic 
tracks the state of the system even in the presence of multiple leaks, although its accuracy gradually deteriorates over time. This allows determining the type of the leakage event: a sharp jump in the value of the statistic indicates an abrupt leak while a gradual increase corresponds to incipient leaks. 

\begin{figure}
    \centering
    \includegraphics[width=0.9\linewidth]{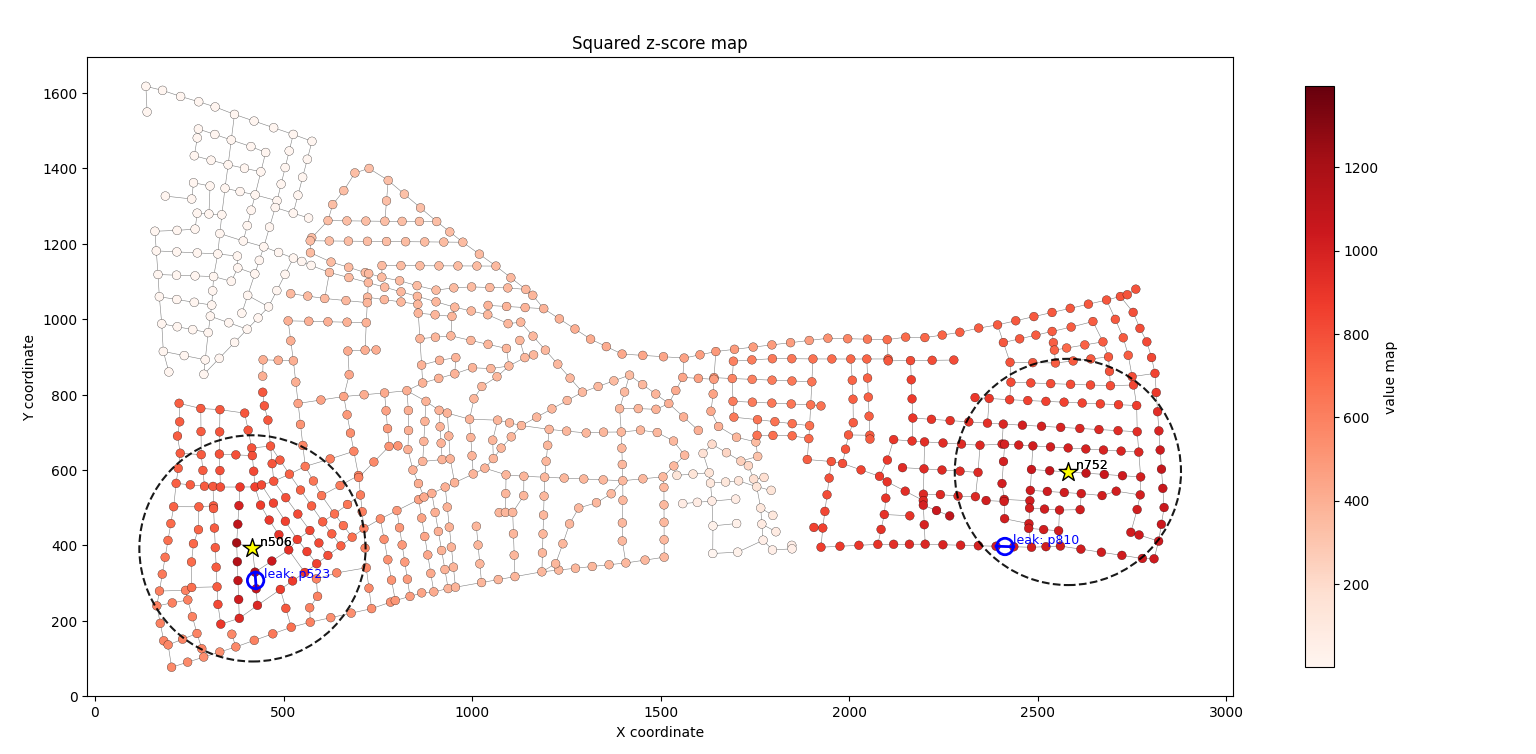}
    \caption{Visual localization of the first two leaks (p523 and p827) in 2019 dataset}
    \label{fig:leaks2019}
\end{figure}

Fig.~\ref{fig:leaks2019} provides an example of visual localization of leakages using a technique described in Sections \ref{sec:classloc}, \ref{sec:localization}.

\begin{figure}[t] 
    \centering
        \begin{subfigure}[c]{0.95\textwidth}
        \includegraphics[width=0.95\linewidth]{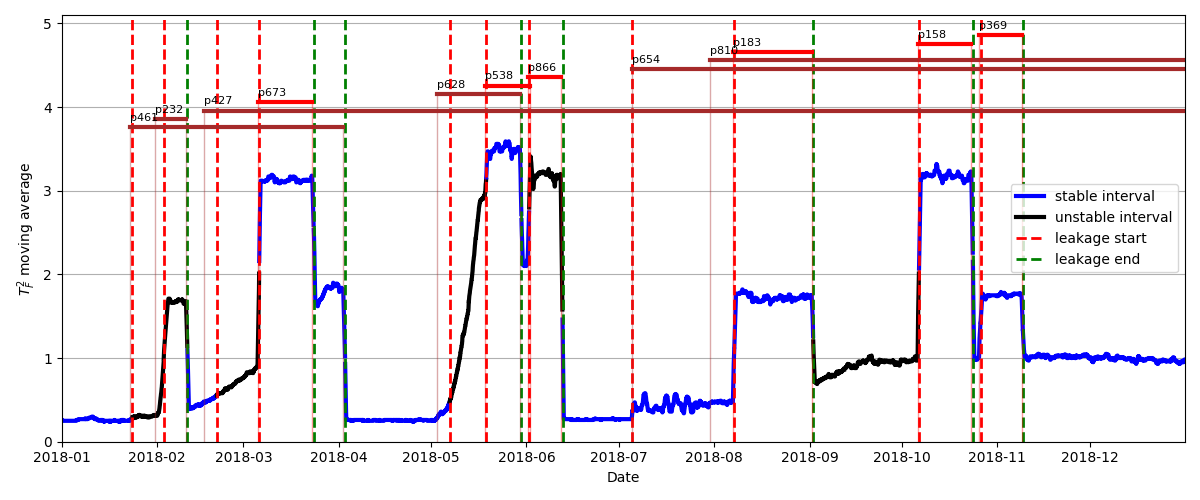}
        \caption{}
        \label{fig:areaA18_PELT}
    \end{subfigure}
    \begin{subfigure}[c]{0.95\textwidth}
        \includegraphics[width=0.95\linewidth]{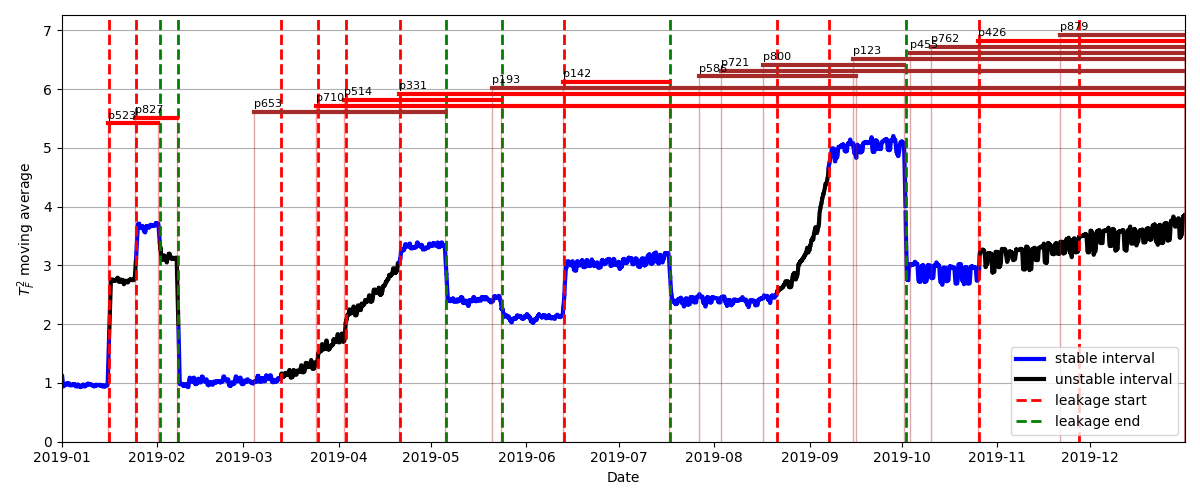}
        \caption{}
        \label{fig:areaA19_PELT}
    \end{subfigure}
    \caption{Leaks detection in area A for 2018 (a) and  2019 (b) using Algorithm 2}
    \label{fig:areaA_PELT} 
\end{figure}

A large number of concurrent leaks in area A makes the application of the hysteresis filter outlined by Algorithm 1 less informative. To recognize leak events, the PELT method and Algorithm 2 were applied to the 12-hour moving average of the $T^2_F$ statistic. The obtained results are illustrated in Fig.~\ref{fig:areaA_PELT}. In this illustration, the commencement of anomalies is indicated in red, and their resolution is denoted in green. Periods of stable operation are depicted in blue, whereas periods of instability are represented in black.

In 2018, out of 12 leak events (6 abrupt and 6 incipient leaks), 11 were successfully detected. Among these, 3 incipient leaks were identified with a delay of 3 to 4 days, while the remaining leaks were detected with a delay of less than half a day.
In 2019, out of 16 leak events (7 abrupt and 9 incipient leaks), 11 were detected (all 7 abrupt and 4 incipient). The incipient leaks were detected with delays ranging from 6 to 21 days, whereas abrupt leaks were identified with a delay of less than half a day.  
However, for leak p426 the localization error exceeded 300 m, thereby classifying the event as a false alarm according to the BattLeDIM criteria.

A complete summary of leak detection results from SICAMS is provided in Table \ref{tab:leak_events} in \ref{sec:app},  where a detected but mislocalized leak p426 is marked with "*". 

\subsection{Performance Metrics}

To evaluate the obtained leak detection results in 2019, we constructed a confusion matrix, following the approach adopted in the final assessment of the BattLeDIM competition. Table \ref{tab:performance} 
compares our results with those 
achieved by the three best-performing teams in BattLeDIM, as well as recent results published in \cite{Barros2025}.

\begin{table}[h]
\centering
\caption{Performance Metrics}
\begin{tabular}{|c|c|c|c|c|}
\hline
Method & \thead{TP rate,\\ \%} & FP count & \thead{Leak-hours\\missed} & \thead{Water lost\\volume, m³} \\ \hline
SICAMS (ours) & 69.57 & 1 & 14,736 & 104,951 \\ \hline
\cite{Barros2025} & 73.91 & 2 & 18,035 & 124,619 \\ \hline
\cite{Stef2022} & 65.22 & 4 & 19,229 & 123,482 \\ \hline
\cite{LiandXin2020} & 56.62 & 3 & - & - \\ \hline
\cite{RomeroBen2022} & 43.47 & 1 & 28,069 & 227,559 \\ \hline
\end{tabular}
\label{tab:performance}
\end{table}

However, the confusion matrix alone does not provide a comprehensive evaluation, as it fails to account for detection delays. 

\begin{figure} [t] 
    \centering
    \includegraphics[width=1\linewidth]{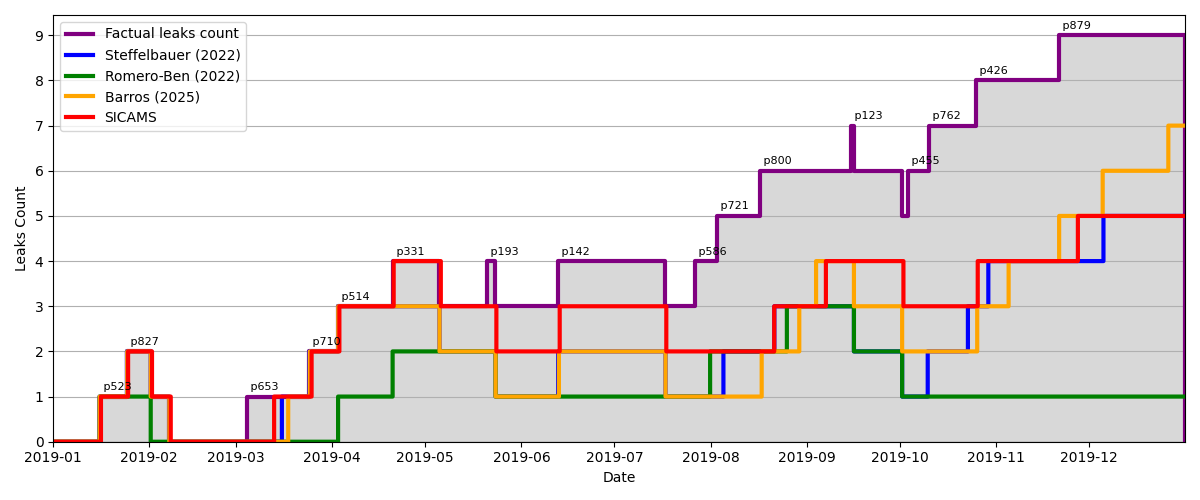}
    \caption{Comparison of detection leaks count in area A in 2019}
    \label{fig:leak_count}
\end{figure}

In accordance with the multiple-detection rule \eqref{eq:mdrule}, we computed the estimates of missed leak-hours by various detection methods using Equation \eqref{eq:metrics}. The computed values of the metric \eqref{eq:metrics} , as well as the estimates of the total volume of lost water, are included in Table \ref{tab:performance}. 
Graphical comparison of the true leak count $\Ntrue{t}$ with its various estimates $\Nest{t}$ are shown in Fig.~\ref{fig:leak_count}. 

\subsection{Computational Costs}
Since model training in SICAMS involves only the computation of mean vectors and covariance matrices for the selected temporal groups and spatial clusters, the overall computational cost is relatively low.
In WNTR-based simulations, a single two-week experiment with a 5-minute time step requires approximately 8.25 s on an Intel i3 @ 3.60 GHz processor with 32 GB of memory. Averaging the results of 100 simulation runs and computing the corresponding whitening matrices takes an additional 6.97 s.

The main portion of this time is spent on hydraulic simulation within WNTR. When processing SCADA data, where the time series are already available, the computation of moments 
grouped by the hour of the day (i.e., 24 mean vectors and covariance matrices) takes less than one second.

It should be noted that averaging is performed only at the training stage, where the mean vectors and covariance matrices are computed over all observations within each temporal cluster. In the evaluation phase, the whitening matrix associated with the cluster (e.g., the hour of the day) is applied to all observations in that cluster, whether the observations are averaged by hour, recorded at 5-minute intervals, or presented in any other temporal resolution.

\section{Extensions}\label{sec:extensions}
\subsection{Estimation of Leakage Volume}
As Hotelling's $T^2_F$ statistic captures multiple leakage events with notable accuracy, it is reasonable to hypothesize a correlation between this statistic and the total water loss in the system. The BattLeDIM dataset provides time series of leakage volumes for each leaking pipe. These data were used to compute the total leakage volume in the network, which was subsequently regressed on the value of the $T^2_F$ statistic. Estimation using the 2018 dataset yields the following models.

a) Linear model:
        \begin{equation}
        \begin{array}{*{20}c}
           L &  =  & {-4.038} &  +  & {15.087} & {F,} & {(R^2  = } & {0.846).}  \\
           {} & {} & {\ (0.110)} & {} & {(0.069)} & {} & {} & {}  \\
        \end{array}
        \label{eq:wlin}
        \end{equation}
 
 b) Log-linear model:
        \begin{equation}
        \begin{array}{*{20}c}
           \ln (1+L) &  =  & { 2.197} &  +  & {1.576} & {\ln F,} & {(R^2  = } & {0.923).}  \\
           {} & {} & {(0.004)} & {} & {(0.005)} & {} & {} & {}  \\
        \end{array}
        \label{eq:wloglin}
        \end{equation}

\begin{figure}[h] 
  \centering
  \begin{subfigure}[c]{0.9\textwidth}
    \includegraphics[width=\textwidth]{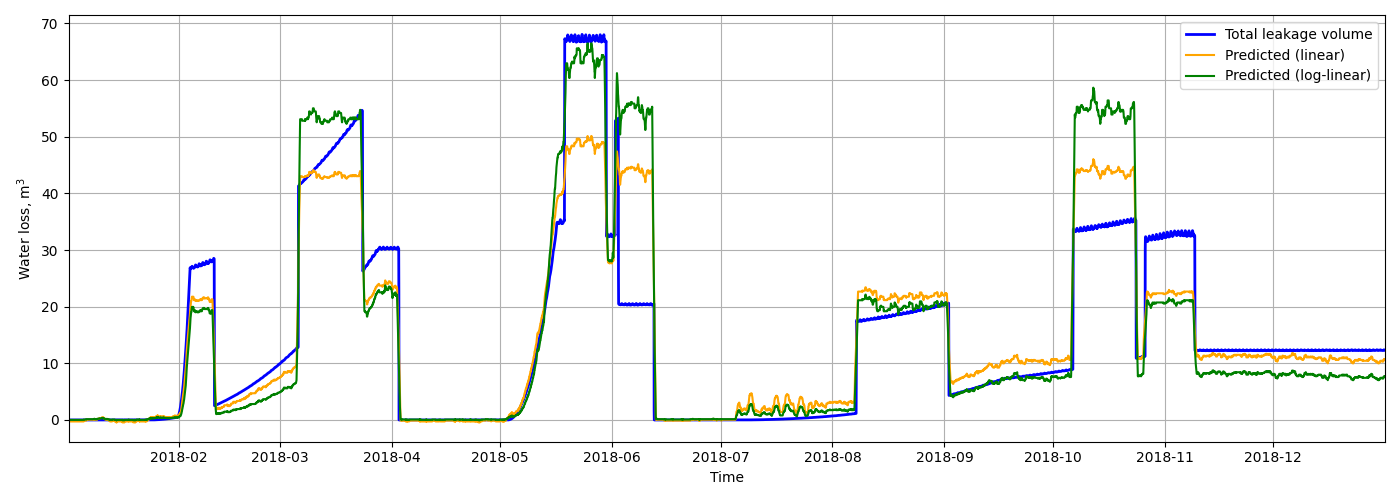}
    \caption{}
    \label{fig:wl18}
  \end{subfigure}
  \hfill
  \begin{subfigure}[c]{0.9\textwidth}
    \includegraphics[width=\textwidth]{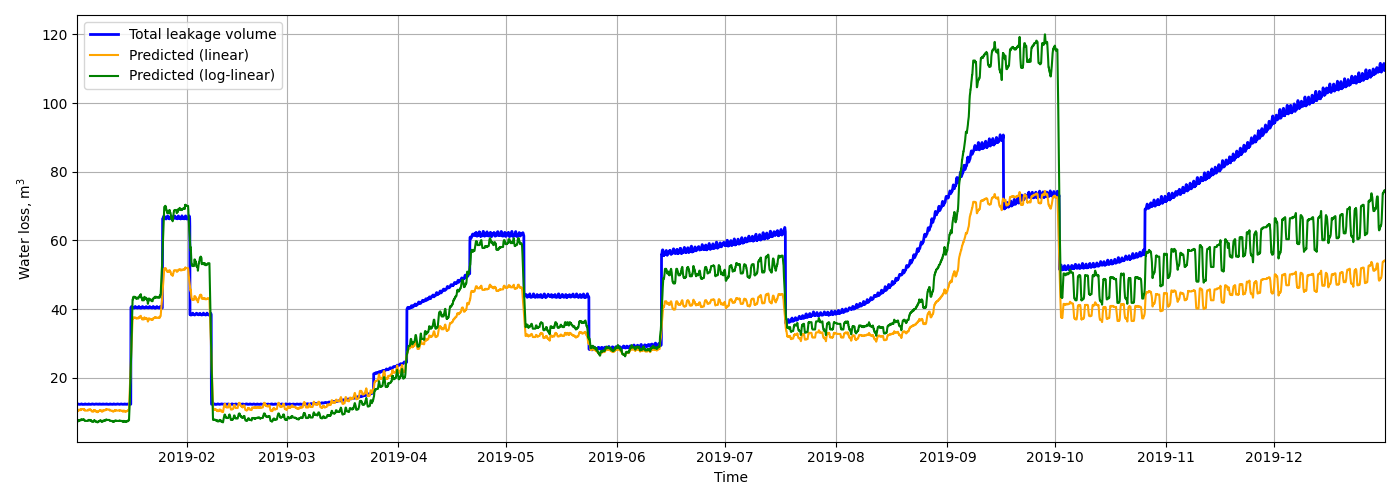}
    \caption{}
    \label{fig:wl19}
  \end{subfigure}
  \caption{Real 
   water loss in 2018 (a) and 2019 (b) SCADA datasets 
   vs. SICAMS estimates}
  \label{fig:wl}
\end{figure}

Here, $L$ denotes the total leakage volume in area A (in m$^3$), excluding leaks p257 and p427, which were considered part of the baseline. $F$ is the 12-period moving average of the $T^2_F$ statistic. Standard errors are reported in parentheses. Fitted values of $L$ for 2018 are shown in Fig.~\ref{fig:wl18},  while Fig.~\ref{fig:wl19} illustrates model predictions 
applied to the 2019 dataset.

The linear model provides a better fit for the 2018 data, whereas the log-linear model performs marginally better for 2019. However, the predictive performance of both models declines over time.

To evaluate the temporal stability of the regression models, the 2019 dataset was divided into two equal half-year periods. Both models demonstrated good agreement with the observed data during the first half of the year, with correlation coefficients between predicted and actual values of 0.968 for the log-linear model and 0.967 for the linear model. However, during the second half of 2019, the model performance deteriorated substantially, with correlations dropping to 0.567 and 0.521, respectively.

This degradation in model adequacy appears to be primarily associated with a significant increase in the total leakage volume. Regression models \eqref{eq:wlin} and \eqref{eq:wloglin} were estimated using the 2018 data, where the total leakage volume only occasionally exceeded 40 m$^3$ (which corresponds to approximately 20\% of the average hourly water consumption in L-Town). In contrast, during the second half of 2019, the total leakage volume never fell below 36 m$^3$ and averaged around 70 m$^3$. Consequently, water loss estimates for this period should be regarded as out-of-sample forecasts, and the decline in predictive accuracy is not unexpected.

\begin{figure}
    \centering
    \includegraphics[width=0.9\linewidth]{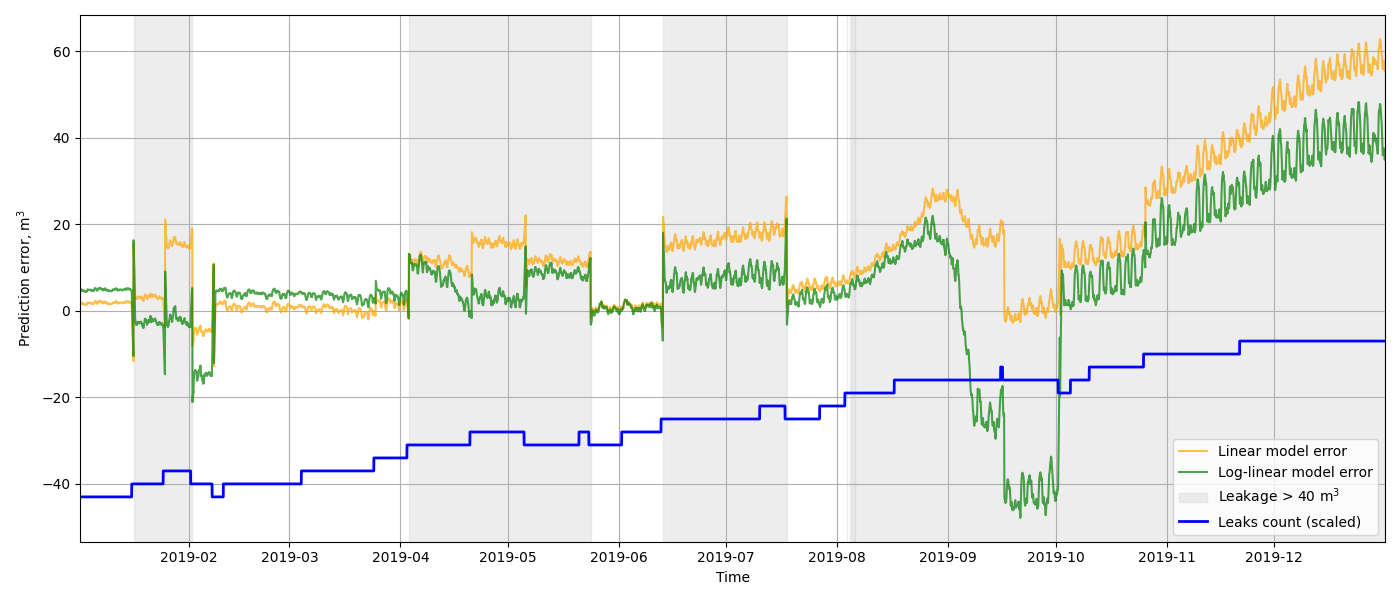}
    \caption{Regression residuals in 2019 as a function of time}
    \label{fig:pred_error}
\end{figure}

This observation is illustrated by Fig.~\ref{fig:pred_error}, which highlights the time intervals when the leakage volume exceeded 40 m$^3$.
The correlation between the residuals and the total leakage volume is 0.864. 
Strong correlation suggests that pooling the 2018 and 2019 datasets could yield a more accurate model. However, this extension was not pursued further, as we tried to maintain comparability with the conditions faced by the BattLeDim participants.

A notable positive correlation was also found between the model error and the total number of leaks in the system (0.773 for the linear model and 0.706 for the log-linear model). This finding suggests that the increasing structural degradation of the system 
also contributes to reduced model accuracy.

Overall, the results indicate that the linear model can provide reliable estimates of the total leakage volume in the network, provided that the deviation from normal operating conditions remains moderate. 


\subsection{Unsupervised Detection}
In practice, it is often challenging for water utilities to obtain a clean dataset that represents fault-free operating conditions. Historical records may be contaminated by unrecorded leak events, changes in consumption patterns, or various sensor faults. As a result, it becomes important to evaluate whether the proposed anomaly detection methodology remains effective in the absence of a dedicated training set.

We compared three strategies for constructing reference (training) statistics in the absence of labeled leak events: 

(i)~dropping out a fixed portion of the lowest-pressure observations for each time cluster. This strategy is motivated by the observation that leaks lead to the overall loss of pressure in the network \citep{SILVA1996S491}; 

(ii)~robust covariance estimation using the Minimum Covariance Determinant (MCD) with an adjustable support fraction \citep{mcd2018}, which drops observations that significantly differ from the bulk of the data;

(iii)~using all available data without filtering.

\begin{figure}
    \centering
    \includegraphics[width=0.9\linewidth]{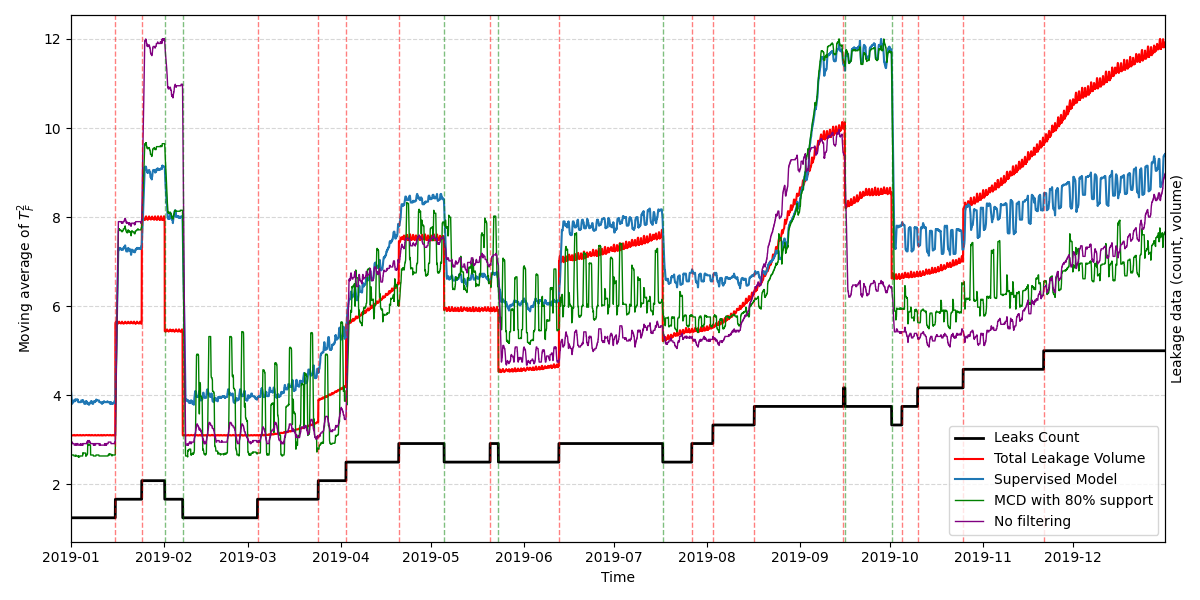}
    \caption{Time series of the $T^2_F$ statistic under different training strategies. Dashed vertical lines mark the onset (red) and termination (green) of leakage events.}
    \label{fig:unsup}
\end{figure}

The first two strategies have yielded similar results.  Somewhat unexpectedly, the third approach has produced results that are, in some ways, superior to the first two methods.
Fig.~\ref{fig:unsup} compares the predictive performance of the $T^2_F$ statistic for Area A under 
different training strategies, after applying a moving average and the Wilson--Hilferty transformation. The true leak count and total leakage volume are shown for reference. All variables are normalized to a common scale.

As illustrated, the absence of a dedicated training set results in a certain degradation of performance. The MCD strategy increases the variance of the test statistic and, consequently, the number of false alarms. Conversely, using an unfiltered training set overly smooths the data, causing some leak events to be missed. Nevertheless, the overall shape of the time series remains similar, and major leakage events are still clearly detectable. This demonstrates that the method 
remains workable with minimal supervision.

One potential enhancement of unsupervised training methods is to add some other unsupervised criteria such as temporal consistency to refine the set of detected outliers. 

\section{Conclusion}
\label{sec:conc}
This study has proposed an integrated framework for anomaly detection in water distribution networks based on multivariate statistical methods
The approach is based on transforming 
sensor data to Mahalanobis space to remove spatial correlations between them. This transformation may be applied to heterogeneous sensor data, enabling seamless integration of measurements from different sensor types. Hotelling's $T^2$ statistic is employed to detect anomalies in real time. 
The method can be used in combination with a hydraulic model or solely through time-series analysis, thus making it feasible in scenarios where network data are incomplete or uncertain.

Our analyses suggest that the $T^2$ statistic can serve as a proxy for overall system condition, which allows the method to remain applicable even under multiple concurrent leaks. Regression models fitted to BattLeDIM data showed that the statistic correlates with the overall leakage volume, facilitating approximate loss estimates that could aid in prioritizing repair tasks.

Beyond detection, we developed a procedure for coarse leak localization using $z$-normalized sensor residuals interpolated over the network graph via the discrete Laplace transformation. This technique produces interpretable spatial fields that highlight likely leak regions. In addition, we designed an event classification algorithm that separates stable and unstable operating periods and distinguishes between abrupt and incipient leakages. 

To address scenarios without fault-free historical data, we evaluated 
several unsupervised training strategies. Despite increased variability and false alarms, major leaks remained clearly detectable, underscoring the robustness of the approach under minimally supervised conditions.

Future work will focus on combining coarse preliminary localization with targeted model-based refinement for precise leak pinpointing. These developments will explore both hydraulic model–assisted and purely data-driven strategies, aiming to deliver scalable and adaptable leak detection tools for operational deployment in modern water distribution systems.

\section*{Acknowledgements}

This work has received support from the Dutch Research Council (NWO) and the National Research Foundation of Ukraine (NRFU), in the scope of the Digital Twin for Evolutionary Changes in Water Networks (DiTEC) project, file numbers 19454 and NWOPI.UKR.2023.001.

\section*{Declaration of generative AI and AI-assisted technologies in the manuscript preparation process}

During the preparation of this work the authors used ChatGPT and Grok in order to improve grammar and style of the text and in order to create code fragments used in the experiments described in the paper. After using these tools/services, the authors reviewed and edited the content as needed and take full responsibility for the content of the published article.

\appendix
\setcounter{table}{0}
\section{Leak Detection Results in the BattLeDIM Dataset}
\label{sec:app}

\begin{table}[H]
\centering
\vspace*{-0.5em}
\caption{Leak Events in L-Town and their Detection Results using SICAMS}
\vspace*{-0.6em} 
\begin{tabular}{|c|c|c|c|c|c|c|c|}
\hline
Pipe & Area & Abrupt & \thead{Max size,\\m³/h} & Start time & \thead{Detection\\ / method} & \thead{Detection\\time} & \thead{$\Delta t$,\\h} \\ \hline
\scalebox{0.9}{p257} & \scalebox{0.9}{C} & \scalebox{0.9}{no} & \scalebox{0.9}{6.87} & \scalebox{0.9}{18-01-08 13:30} & \scalebox{0.9}{yes/filter} & \scalebox{0.9}{18-01-18 13:00} & \scalebox{0.9}{239} \\ \hline
\scalebox{0.9}{p461} & \scalebox{0.9}{A} & \scalebox{0.9}{no} & \scalebox{0.9}{30.54} & \scalebox{0.9}{18-01-23 04:25} & \scalebox{0.9}{yes/PELT} & \scalebox{0.9}{18-01-23 22:00} & \scalebox{0.9}{17} \\ \hline
\scalebox{0.9}{p232} & \scalebox{0.9}{A} & \scalebox{0.9}{no} & \scalebox{0.9}{26.1} & \scalebox{0.9}{18-01-31 03:45} & \scalebox{0.9}{yes/PELT} & \scalebox{0.9}{18-02-03 03:00} & \scalebox{0.9}{71} \\ \hline
\scalebox{0.9}{p427} & \scalebox{0.9}{A} & \scalebox{0.9}{no} & \scalebox{0.9}{5.11} & \scalebox{0.9}{18-02-16 04:15} & \scalebox{0.9}{yes/PELT} & \scalebox{0.9}{18-02-20 10:00} & \scalebox{0.9}{101} \\ \hline
\scalebox{0.9}{p673} & \scalebox{0.9}{B} & \scalebox{0.9}{yes} & \scalebox{0.9}{28.42} & \scalebox{0.9}{18-03-05 15:45} & \scalebox{0.9}{yes/filter} & \scalebox{0.9}{18-03-05 18:00} & \scalebox{0.9}{2} \\ \hline
\scalebox{0.9}{p628} & \scalebox{0.9}{A} & \scalebox{0.9}{no} & \scalebox{0.9}{35.38} & \scalebox{0.9}{18-05-02 18:50} & \scalebox{0.9}{yes/PELT} & \scalebox{0.9}{18-05-07 01:00} & \scalebox{0.9}{102} \\ \hline
\scalebox{0.9}{p538} & \scalebox{0.9}{A} & \scalebox{0.9}{yes} & \scalebox{0.9}{32.84} & \scalebox{0.9}{18-05-18 08:35} & \scalebox{0.9}{yes/PELT} & \scalebox{0.9}{18-05-18 22:00} & \scalebox{0.9}{13} \\ \hline
\scalebox{0.9}{p866} & \scalebox{0.9}{A} & \scalebox{0.9}{yes} & \scalebox{0.9}{20.57} & \scalebox{0.9}{18-06-01 09:05} & \scalebox{0.9}{yes/PELT} & \scalebox{0.9}{18-06-01 21:00} & \scalebox{0.9}{11} \\ \hline
\scalebox{0.9}{p31} & \scalebox{0.9}{C} & \scalebox{0.9}{no} & \scalebox{0.9}{16.21} & \scalebox{0.9}{18-06-29 01:35} & \scalebox{0.9}{yes/filter} &  \scalebox{0.9}{18-07-05 8:00} &  \scalebox{0.9}{149} \\ \hline
\scalebox{0.9}{p654} & \scalebox{0.9}{A} & \scalebox{0.9}{no} & \scalebox{0.9}{5.49} & \scalebox{0.9}{18-07-05 03:40} & \scalebox{0.9}{yes/PELT} & \scalebox{0.9}{18-07-05 10:00} & \scalebox{0.9}{6} \\ \hline
\scalebox{0.9}{p810} & \scalebox{0.9}{A} & \scalebox{0.9}{no} & \scalebox{0.9}{6.91} & \scalebox{0.9}{18-07-30 18:40} & \scalebox{0.9}{no} & \scalebox{0.9}{-} & \scalebox{0.9}{-} \\ \hline
\scalebox{0.9}{p183} & \scalebox{0.9}{A} & \scalebox{0.9}{yes} & \scalebox{0.9}{16.46} & \scalebox{0.9}{18-08-07 02:35} & \scalebox{0.9}{yes/PELT} & \scalebox{0.9}{18-08-07 13:00} & \scalebox{0.9}{10} \\ \hline
\scalebox{0.9}{p158} & \scalebox{0.9}{A} & \scalebox{0.9}{yes} & \scalebox{0.9}{24.76} & \scalebox{0.9}{18-10-06 02:35} & \scalebox{0.9}{yes/PELT} & \scalebox{0.9}{18-10-06 13:00} & \scalebox{0.9}{10} \\ \hline
\scalebox{0.9}{p369} & \scalebox{0.9}{A} & \scalebox{0.9}{yes} & \scalebox{0.9}{21.05} & \scalebox{0.9}{18-10-26 02:05} & \scalebox{0.9}{yes/PELT} & \scalebox{0.9}{18-10-26 13:00} & \scalebox{0.9}{11} \\ \hline
\scalebox{0.9}{p523} & \scalebox{0.9}{A} & \scalebox{0.9}{yes} & \scalebox{0.9}{28.39} & \scalebox{0.9}{19-01-15 23:00} & \scalebox{0.9}{yes/PELT} & \scalebox{0.9}{19-01-16 10:00} & \scalebox{0.9}{11} \\ \hline
\scalebox{0.9}{p827} & \scalebox{0.9}{A} & \scalebox{0.9}{yes} & \scalebox{0.9}{26.46} & \scalebox{0.9}{19-01-24 18:30} & \scalebox{0.9}{yes/PELT} & \scalebox{0.9}{19-01-25 04:00} & \scalebox{0.9}{9} \\ \hline
\scalebox{0.9}{p280} & \scalebox{0.9}{C} & \scalebox{0.9}{yes} & \scalebox{0.9}{5.26} & \scalebox{0.9}{19-02-10 13:05} & \scalebox{0.9}{yes/filter} & \scalebox{0.9}{19-02-11 9:00} & \scalebox{0.9}{19} \\ \hline
\scalebox{0.9}{p653} & \scalebox{0.9}{A} & \scalebox{0.9}{no} & \scalebox{0.9}{18.28} & \scalebox{0.9}{19-03-04 08:40} & \scalebox{0.9}{yes/PELT} & \scalebox{0.9}{19-03-13 06:00} & \scalebox{0.9}{213} \\ \hline
\scalebox{0.9}{p710} & \scalebox{0.9}{A} & \scalebox{0.9}{yes} & \scalebox{0.9}{5.58} & \scalebox{0.9}{19-03-24 14:15} & \scalebox{0.9}{yes/PELT} & \scalebox{0.9}{19-03-25 08:00} & \scalebox{0.9}{17} \\ \hline
\scalebox{0.9}{p514} & \scalebox{0.9}{A} & \scalebox{0.9}{yes} & \scalebox{0.9}{15.58} & \scalebox{0.9}{19-04-02 20:40} & \scalebox{0.9}{yes/PELT} & \scalebox{0.9}{19-04-03 07:00} & \scalebox{0.9}{10} \\ \hline
\scalebox{0.9}{p331} & \scalebox{0.9}{A} & \scalebox{0.9}{yes} & \scalebox{0.9}{10.93} & \scalebox{0.9}{19-04-20 10:10} & \scalebox{0.9}{yes/PELT} & \scalebox{0.9}{19-04-20 19:00} & \scalebox{0.9}{8} \\ \hline
\scalebox{0.9}{p193} & \scalebox{0.9}{A} & \scalebox{0.9}{no} & \scalebox{0.9}{10.36} & \scalebox{0.9}{19-05-20 21:55} & \scalebox{0.9}{no} & \scalebox{0.9}{-} & \scalebox{0.9}{-} \\ \hline
\scalebox{0.9}{p277} & \scalebox{0.9}{C} & \scalebox{0.9}{no} & \scalebox{0.9}{7.36} & \scalebox{0.9}{19-06-01 19:40} & \scalebox{0.9}{yes/filter} & \scalebox{0.9}{19-06-20 04:00} & \scalebox{0.9}{439} \\ \hline
\scalebox{0.9}{p142} & \scalebox{0.9}{A} & \scalebox{0.9}{yes} & \scalebox{0.9}{27.04} & \scalebox{0.9}{19-06-12 19:55} & \scalebox{0.9}{yes/PELT} & \scalebox{0.9}{19-06-13 08:00} & \scalebox{0.9}{12} \\ \hline
\scalebox{0.9}{p680} & \scalebox{0.9}{B} & \scalebox{0.9}{no} & \scalebox{0.9}{5.37} & \scalebox{0.9}{19-07-10 08:45} & \scalebox{0.9}{yes/filter} & \scalebox{0.9}{19-06-13 11:00} & \scalebox{0.9}{2} \\ \hline
\scalebox{0.9}{p586} & \scalebox{0.9}{A} & \scalebox{0.9}{no} & \scalebox{0.9}{20.52} & \scalebox{0.9}{19-07-27 02:55} & \scalebox{0.9}{no} & \scalebox{0.9}{-} & \scalebox{0.9}{-} \\ \hline
\scalebox{0.9}{p721} & \scalebox{0.9}{A} & \scalebox{0.9}{no} & \scalebox{0.9}{13.18} & \scalebox{0.9}{19-08-03 03:20} & \scalebox{0.9}{yes/PELT} & \scalebox{0.9}{19-08-21 12:00} & \scalebox{0.9}{440} \\ \hline
\scalebox{0.9}{p800} & \scalebox{0.9}{A} & \scalebox{0.9}{no} & \scalebox{0.9}{21.95} & \scalebox{0.9}{19-08-16 22:05} & \scalebox{0.9}{yes/PELT} & \scalebox{0.9}{19-09-07 04:00} & \scalebox{0.9}{510} \\ \hline
\scalebox{0.9}{p123} & \scalebox{0.9}{A} & \scalebox{0.9}{no} & \scalebox{0.9}{9.19} & \scalebox{0.9}{19-09-15 15:40} & \scalebox{0.9}{no} & \scalebox{0.9}{-} & \scalebox{0.9}{-} \\ \hline
\scalebox{0.9}{p455} & \scalebox{0.9}{A} & \scalebox{0.9}{no} & \scalebox{0.9}{11.05} & \scalebox{0.9}{19-10-05 03:35} & \scalebox{0.9}{no} & \scalebox{0.9}{-} & \scalebox{0.9}{-} \\ \hline
\scalebox{0.9}{p762} & \scalebox{0.9}{A} & \scalebox{0.9}{no} & \scalebox{0.9}{15.71} & \scalebox{0.9}{19-10-10 09:40} & \scalebox{0.9}{no} & \scalebox{0.9}{-} & \scalebox{0.9}{-} \\ \hline
\scalebox{0.9}{p426*} & \scalebox{0.9}{A} & \scalebox{0.9}{yes} & \scalebox{0.9}{13.56} & \scalebox{0.9}{19-10-25 13:25} & \scalebox{0.9}{yes/PELT} & \scalebox{0.9}{19-10-26 03:00} & \scalebox{0.9}{13} \\ \hline
\scalebox{0.9}{p879} & \scalebox{0.9}{A} & \scalebox{0.9}{no} & \scalebox{0.9}{10.93} & \scalebox{0.9}{19-11-21 09:15} & \scalebox{0.9}{yes/PELT} & \scalebox{0.9}{19-11-27 10:00} & \scalebox{0.9}{144} \\ \hline
\end{tabular}
\label{tab:leak_events}
\end{table}

\bibliographystyle{elsarticle-harv} 
\bibliography{refs}

@ARTICLE{LW19,
  author  = {Liemberger, R. and Wyatt, A.},
  title   = {Quantifying the global non-revenue water problem},
  journal = {Water Supply}, 
  volume  = {19},
  number = {3},
  year   = {2019},
  pages  = {831--837},
  doi = {10.2166/ws.2018.129}
}

@ARTICLE{Vrachimis2022,
  author  = {Vrachimis, S. G. and Eliades, D. G. and Taormina, R. and Kapelan, Z. and Ostfeld, A. and Liu, S. and Kyriakou, M. and Pavlou, P. and Qiu, M. and Polycarpou, M. M.},
  title   = {Battle of the {Leakage} {Detection} and {Isolation} {Methods}},
  journal = {J. Water Res. Plann. Manage.}, 
  volume  = {148},
  number = {12},
  year   = {2022},
  pages  = {04022068},
  doi = {10.1061/(ASCE)WR.1943-5452.0001601}
}

@ARTICLE{Hu2021,
  author  = {Hu, Z. and Chen, B. and Chen, W. and Tan, D. and Shen, D.},
  title   = {Review of model-based and data-driven approaches for leak detection and location in water distribution systems},
  journal = {Water Supply}, 
  volume  = {21},
  number = {7},
  year   = {2021},
  pages  = {3282–-3306},
  doi = {10.2166/ws.2021.101}
}

@ARTICLE{Yussof2021,
  author  = {Mohd Yussof, N. A. and Ho, H. W.},
  title   = {Review of water leak detection methods in smart building applications},
  journal = {Buildings}, 
  volume  = {12},
  number = {10},
  year   = {2022},
  pages  = {1535},
  doi = {10.3390/buildings12101535}
}

@CONFERENCE{Negm2023,
  author  = {Negm, A. and Ma, X. and Aggidis, G.},
  title   = {Review of leakage detection in water distribution networks},
  booktitle = {In Proc. 14th Int. Conf. on Hydroinformatics}, 
  year    = {2023},
  volume  = {1136},
  pages   = {012052},
  doi = {10.1088/1755-1315/1136/1/012052}
}

@ARTICLE{Romero2023,
  author  = {Romero-Ben, L. and Alves, D. and Blesa, J. and Cembrano, G. and Puig, V. and Duviella, E.},
  title   = {Leak detection and localization in water distribution networks: Review and perspective},
  journal = {Annual Reviews in Control}, 
  volume  = {55},
  year   = {2023},
  pages  = {392--419},
  doi = {10.1016/j.arcontrol.2023.03.012}
}

@ARTICLE{Wan2022,
  author  = {Wan, X. and Kuhanestani, P. K. and Farmani, R. and Keedwell, E.},
  title   = {Literature review of data analytics for leak detection in water distribution networks: A focus on pressure and flow smart sensors},
  journal = {J. Water Resour. Plann. Manage.}, 
  volume  = {148},
  pages = {10},
  year   = {2022},
  doi = {10.1061/(ASCE)WR.1943-5452.0001597}
}

@ARTICLE{Loureiro2016,
  author  = {Loureiro, D. and Amado, C. and Martins, A. and Vitorino, D. and Mamade, A. and Coelho, S.T.},
  title   = {Water distribution systems flow monitoring and anomalous event detection: A practical approach},
  journal = {Urban Water Journal}, 
  volume  = {13},
  number = {3},
  pages = {242--252},
  year   = {2016}
}

@ARTICLE{Ahn2019,
  author  = {Ahn, J. and Jung, D.},
  title   = {Hybrid statistical process control method for water distribution pipe burst detection},
  journal = {J. Water Resour. Plann. Manage.}, 
  volume  = {145},
  number = {9},
  pages = {06019008},
  year   = {2019}
}

@ARTICLE{Jung2015,
  author  = {Jung, D. and Kang, D. and Liu, J. and Lansey, K.},
  title   = {Improving the rapidity of responses to pipe burst in water distribution systems: A comparison of statistical process control methods},
  journal = {Journal of Hydroinformatics}, 
  volume  = {17},
  number = {2},
  pages = {307--328},
  year   = {2015}
}

@ARTICLE{ChenYang2022,
  author  = {Chen, Y. and Yang, Y.},
  title   = {A multivariate statistical model of water leakage in urban water supply networks based on random matrix theory},
  journal = {Mathematical Problems in Engineering}, 
  pages = {1--11.},
  year   = {2022},
  doi = {10.1155/2022/2314972}
}

@ARTICLE{Xie2024,
  author  = {Xie, C. and Tian, Z. and Chen, J. and Fang, Q. and Wang, J.},
  title   = {A multi-point leakage prediction statistical method using {Bayesian} inference in water distribution networks},
  journal = {Water Practice and Technology}, 
  volume  = {19},
  number = {10},
  pages = {3987--4000},
  year   = {2024},
  doi = {10.2166/wpt.2024.233}
}

@ARTICLE{CaiGaoXu2022,
  author  = {Cai, J. and Gao, J. and Xu, Y.},
  title   = {Anomaly detection and classification in water distribution networks integrated with hourly nodal water demand forecasting models and feature extraction technique},
  journal = {J. Water Resour. Plann. Manage.}, 
  volume  = {148},
  number = {11},
  year   = {2022},
  doi = {10.1061/(asce)wr.1943-5452.0001616}
}

@ARTICLE{Mohan2023,
  author  = {Mohan Doss, P. and Rokstad, M. M. and Steffelbauer, D. and Tscheikner-Gratl, F.},
  title   = {Uncertainties in different leak localization methods for water distribution networks: a review},
  journal = {Urban Water Journal}, 
  volume  = {20},
  number = {8},
  year   = {2023},
  pages  = {953--967},
  doi = {10.1080/1573062X.2023.2229301}
}

@ARTICLE{Nimri2023,
  author  = {Nimri, W. and Wang, Y. and Zhang, Z. and Deng, C. and Sellstrom, K.},
  title   = {Data-driven approaches and model-based methods for detecting and locating leaks in water distribution systems: a literature review},
  journal = {Neural Computing and Applications}, 
  volume  = {35},
  year   = {2023},
  pages  = {11611--11623},
  doi = {10.1007/s00521-023-08497-x}
}

@ARTICLE{Farah2023,
  author  = {Farah, E. and Shahrour, I.},
  title   = {Water leak detection: A comprehensive review of methods, challenges, and future directions},
  journal = {Water}, 
  volume  = {16},
  year   = {2023},
  pages  = {2975},
  doi = {10.3390/w16202975}
}

@ARTICLE{Liu2024,
  author  = {Liu, R. and Zayed, T. and Xiao, R.},
  title   = {Advanced acoustic leak detection in water distribution networks using integrated generative mode},
  journal = {Water Research}, 
  volume  = {254},
  year    = {2024},
  pages  = {121434},
  doi = {10.1016/j.watres.2024.121434}
}

@ARTICLE{Min2022,
  author  = {Min, K. W. and Kim, T. and Lee, S. and Choi, Y. H. and Kim, J. H.},
  title   = {Detecting and localizing leakages in water distribution systems using a two-phase model},
  journal = {J. Water Resour. Plann. Manage.}, 
  volume  = {148},
  number = {10},
  year    = {2022},
  pages  = {0402205},
  doi = {10.1061/(ASCE)WR.1943-5452.0001599}
}

@ARTICLE{Barros2025,
  author  = {Barros, D. and Zanfei, A. and Menapace, A. and Meirelles, G. and Herrera, M. and Brentan, B.},
  title   = {Leak detection and localization in water distribution systems via multilayer networks},
  journal = {Water Research X}, 
  volume  = {26},
  year    = {2025},
  pages  = {100280},
  doi = {10.1016/j.wroa.2024.100280}
}

@ARTICLE{Shen2022,
  author  = {Shen, Y. and Cheng, W.},
  title   = {A tree-based machine learning method for pipeline leakage detection},
  journal = {Water}, 
  volume  = {14},
  number = {18},
  year    = {2022},
  pages  = {2833},
  doi = {10.3390/w14182833}
}

@ARTICLE{Marzola2022,
  author  = {Marzola, I. and Alvisi, S. and Franchini, M.},
  title   = {A comparison of model-based methods for leakage localization in water distribution systems},
  journal = {Int. J. EWRA}, 
  volume  = {36},
  number = {14},
  year    = {2022},
  pages  = {5711--5727},
  doi = {10.1007/s11269-022-03329-4}
}

@CONFERENCE{Cai2022,
  author  = {Cai, Z. and Dziedzic, R. and Li, S.S.},
  title   = {Water distribution system leak detection using {Support} {Vector} {Machines}},
  booktitle = {Proc. CSCE 2021. Lecture Notes in Civil Engineering}, 
  volume  = {250},
  year    = {2022},
  doi = {10.1007/978-981-19-1065-4\_41}
}

@ARTICLE{Truong2024,
  author  = {Truong, H. and Tello, A. and Lazovik, A. and Degeler, V.},
  title   = {Graph neural networks for pressure estimation in water distribution systems},
  journal = {Water Resour. Res.}, 
  volume  = {60},
  year    = {2024},
  doi = {10.1029/2023wr036741}
}

@ARTICLE{Daniel2022,
  author  = {Daniel, I. and Pesantez, J. and Letzgus, S. and Khaksar Fasaee, M. A. and Alghamdi, F. and Berglund, E. and Mahinthakumar, G. and Cominola, A.},
  title   = {A sequential pressure-based algorithm for data-driven leakage identification and model-based localization in water distribution networks},
  journal = {J. Water Resour. Plann. Manage.}, 
  volume  = {148},
  number = {6},
  year    = {2022},
  pages  = {04022025},
  doi = {10.1061/(ASCE)WR.1943-5452.0001535}
}

@ARTICLE{Mazzoni2024,
  author  = {Mazzoni, F. and Marsili, V. and Alvisi, S. and Franchini, M.},
  title   = {Detection and pre-localization of anomalous consumption events in water distribution networks through automated, pressure-based methodology},
  journal = {Water Resour. Indust.}, 
  volume  = {31},
  year    = {2024},
  pages  = {100255},
  doi = {10.1016/j.wri.2024.100255}
}

@book{mva2015,
  title     = "Applied multivariate statistical analysis",
  author    = "Härdle, W. and Simar, L.",
  year      = {2015},
  publisher = "Berlin, Heidelberg: Springer"
}

@ARTICLE{whitening2018,
  author  = {Kessy, A. and Lewin, A. and Strimmer, K.},
  title   = {Optimal whitening and decorrelation},
  journal = {The American Statistician}, 
  volume  = {72},
  number = {4},
  year    = {2018},
  pages  = {309--314}
}

@ARTICLE{Bui2020,
  author  = {Khoa Bui, X. and Marlim, M. S. and Kang, D.},
  title   = {Water network partitioning into {District} {Metered} {Areas}: A state-of-the-art review},
  journal = {Water}, 
  volume  = {12},
  number = {4},
  year    = {2020},
  pages  = {1002}
}

@ARTICLE{wh1931,
  author  = {Wilson, E. B. and Hilferty, M. M.},
  title   = {The distribution of chi-square},
  journal = {Proceedings of the National Academy of Sciences}, 
  volume  = {17},
  number = {12},
  year    = {1931},
  pages  = {684--688}
}

@ARTICLE{pelt2012,
  author  = {Killick, R. and Fearnhead, P. and Eckley, I.A.},
  title   = {Optimal detection of changepoints with a linear computational cost},
  journal = { Journal of the American Statistical Association}, 
  volume  = {107},
  number = {500},
  year    = {2012},
  pages  = {1590--1598}
}

@ARTICLE{mcd2018,
  author  = {Hubert, M. and Debruyne, M. and Rousseeuw, P.J.},
  title   = {Minimum covariance determinant and extensions},
  journal = {Wiley Interdisciplinary Reviews: Computational Statistics}, 
  volume  = {10},
  number = {53},
  year    = {2018},
  pages  = {e1421}
}

@ARTICLE{Stef2022,
  author  = {Steffelbauer, D. B. and Deuerlein, J. and Gilbert, D. and Abraham, E. and Piller, O.},
  title   = {Pressure-leak duality for leak detection and localization in water distribution systems},
  journal = {J. Water Resour. Plann. Manage.}, 
  volume  = {148},
  number = {3},
  year    = {2022},
  pages  = {04021106},
  doi = {10.1061/(ASCE)WR.1943-5452.0001515}
}

@ARTICLE{Wang2022,
  author  = {Wang, X. and Li, J. and Liu, S. and Yu, X. and Ma, Z.},
  title   = {Multiple leakage detection and isolation in district metering areas using a multistage approach},
  journal = {J. Water Resour. Plann. Manage.}, 
  volume  = {148},
  number = {6},
  year    = {2022},
  pages  = {04022021},
  doi = {10.1061/(ASCE)WR.1943-5452.0001558}
}

@ARTICLE{LiandXin2020,
  author  = {Li, Z. and Xin, K.},
  title   = {Fast localization of multiple leaks in water distribution
network jointly driven by simulation and machine learning},
  journal = {Geneva: CERN Data Center.}, 
  year    = {2020},
  doi = {10.5281/zenodo.3911045}
}

@ARTICLE{RomeroBen2022,
  author  = {Romero-Ben, L. and Alves, D. and Blesa, J. and Cembrano, G. and Puig, V.},
  title   = {Leak localization in water distribution networks using data-driven and model-based approaches},
  journal = {J. Water Resour. Plann. Manage.}, 
  volume  = {148},
  number = {5},
  year    = {2022},
  doi = {10.1061/(ASCE)WR.1943-5452.0001542}
}

@techreport{Rossman2000,
  author       = {Lewis A. Rossman},
  title        = {EPANET 2: Users Manual},
  institution  = {United States Environmental Protection Agency},
  year         = {2000},
  address      = {Cincinnati, OH, USA}
}

@article{Preis2011,
  author  = {Preis, A. and Whittle, A. J. and Ostfeld, A. and Janke, R.},
  title   = {On-line hydraulic state estimation in urban water networks using the {Kalman} filter},
  journal = {J. Water Resour. Plann. Manage.},
  year    = {2011},
  volume  = {137},
  number  = {4},
  pages   = {343--351},
  doi     = {10.1061/(ASCE)WR.1943-5452.0000124}
}

@InProceedings{ghorbani19,
  title = 	 {Data {Shapley}: Equitable Valuation of Data for Machine Learning},
  author =       {Ghorbani, Amirata and Zou, James},
  booktitle = 	 {Proceedings of the 36th International Conference on Machine Learning},
  pages = 	 {2242--2251},
  year = 	 {2019},
  editor = 	 {Chaudhuri, Kamalika and Salakhutdinov, Ruslan},
  volume = 	 {97},
  series = 	 {Proceedings of Machine Learning Research},
  month = 	 {09--15 Jun},
  url = 	 {https://proceedings.mlr.press/v97/ghorbani19c.html},
}

@article{jian2022,
  author = {Jian, C. and Gao, J. and Xu, Y.},
  title = {Anomaly detection and classification in water distribution networks integrated with hourly nodal water demand forecasting models and feature extraction technique},
  journal = {J. Water Resour. Plann. Manage.},
  volume = {148},
  number = {11},
  pages = {04022059},
  year = {2022}
}

@article{SILVA1996S491,
title = {Pressure wave behaviour and leak detection in pipelines},
journal = {Comput. Chem. Eng.},
volume = {20},
pages = {S491--S496},
year = {1996},
doi = {https://doi.org/10.1016/0098-1354(96)00091-9},
author = {Reinaldo A. Silva and Claudio M. Buiatti and Sandra L. Cruz and João A.F.R. Pereira},
}

@article{MahExample1,
title = {Anomaly detection of cooling fan and fault classification of induction motor using {Mahalanobis}–{Taguchi} system},
journal = {Expert Systems with Applications},
volume = {40},
number = {15},
pages = {5787--5795},
year = {2013},
doi = {10.1016/j.eswa.2013.04.024},
author = {Xiaohang Jin and Tommy W.S. Chow},
}

@article{MahExample2,
title = {An effective approach to detect the source(s) of out-of-control signals in productive processes by vector error correction (VEC) residual and {Hotelling’s} ${T}^2$ decomposition techniques},
journal = {Expert Systems with Applications},
volume = {187},
pages = {115979},
year = {2022},
doi = {10.1016/j.eswa.2021.115979},
author = {Renan Mitsuo Ueda and Adriano Mendonça Souza},
}

@article{Clements2024,
title = {Survey expectations and adjustments for multiple testing},
journal = {Journal of Economic Behavior \& Organization},
volume = {224},
pages = {338--354},
year = {2024},
doi = {10.1016/j.jebo.2024.06.009},
author = {Michael P. Clements},
}

@book{american2008water,
  title={Water audits and loss control programs: M36},
  author={AWWA},
  volume={36},
  year={2008},
  publisher={American Water Works Association}
}

@article{palau2012burst,
  title={Burst detection in water networks using principal component analysis},
  author={Palau, CV and Arregui, FJ and Carlos, M},
  journal={Journal of Water Resources Planning and Management},
  volume={138},
  number={1},
  pages={47--54},
  year={2012},
}

\end{document}